\pdfoutput=1
\PassOptionsToPackage{numbers}{natbib}

\documentclass{article}

\usepackage[final]{nips}

\usepackage{microtype}
\usepackage{graphicx}
\usepackage{subfigure}
\usepackage{booktabs}

\usepackage{xcolor}
\usepackage{hyperref}
\hypersetup{
    colorlinks,
    linkcolor={blue!50!black},
    citecolor={blue!50!black},
    urlcolor={blue!50!black}
}
\usepackage{url}
\usepackage[utf8]{inputenc} 
\usepackage[T1]{fontenc}    
\usepackage{amsfonts}       
\usepackage{nicefrac}       
\usepackage[titletoc,title]{appendix}
\usepackage{graphics}
\usepackage{float}
\usepackage{amsmath}
\usepackage[stable]{footmisc}
\newcommand{\bepsilon}{\boldsymbol{\epsilon}}
\newcommand{\btheta}{\boldsymbol{\theta}}
\newcommand{\bphi}{\boldsymbol{\phi}}
\newcommand{\myeq}[1]{\hyperref[eq:#1]{Equation~\ref*{eq:#1}}}
\newcommand{\myeqq}[2]{\hyperref[eq:#1]{Equation~\ref*{eq:#1}},\hyperref[eq:#2]{\ref*{eq:#2}}}
\newcommand{\mysub}[1]{\hyperref[sub:#1]{Section~\ref*{sub:#1}}}
\newcommand{\mysec}[1]{\hyperref[sec:#1]{Section~\ref*{sec:#1}}}
\newcommand{\mychapter}[1]{\hyperref[chapter:#1]{Chapter~\ref*{chapter:#1}}}
\newcommand{\mytable}[1]{\hyperref[table:#1]{Table~\ref*{table:#1}}}
\newcommand{\myfig}[1]{\hyperref[fig:#1]{Figure~\ref*{fig:#1}}}
\newcommand{\myfigg}[2]{\hyperref[fig:#1]{Figure~\ref*{fig:#1}#2}}
\newcommand{\myfiggg}[3]{\hyperref[fig:#1]{Figure~\ref*{fig:#1}#2,#3}}
\newcommand{\myappendix}[1]{\hyperref[appendix:#1]{Appendix~\ref*{appendix:#1}}}
\newcommand{\myalg}[1]{\hyperref[alg:#1]{Algorithm~\ref*{alg:#1}}}
\newcommand{\mytheorem}[1]{\hyperref[theorem:#1]{Theorem~\ref*{theorem:#1}}}
\newcommand{\myfootnote}[1]{\hyperref[footnote:#1]{Footnote~\ref*{footnote:#1}}}
\DeclareRobustCommand{\parhead}[1]{\textbf{#1}~}

\author{
Alireza Makhzani\\
Vector Institute for Artificial Intelligence\\University of Toronto\\
\texttt{makhzani@vectorinstitute.ai}
}

\title{Implicit Autoencoders}

\begin{document}

\maketitle

\begin{abstract}
In this paper, we describe the ``implicit autoencoder'' (IAE), a generative autoencoder in which both the generative path and the recognition path are parametrized by implicit distributions. We use two generative adversarial networks to define the reconstruction and the regularization cost functions of the implicit autoencoder, and derive the learning rules based on maximum-likelihood learning. Using implicit distributions allows us to learn more expressive posterior and conditional likelihood distributions for the autoencoder. Learning an expressive conditional likelihood distribution enables the latent code to only capture the abstract and high-level information of the data, while the remaining low-level information is captured by the implicit conditional likelihood distribution. We show the applications of implicit autoencoders in disentangling content and style information, clustering, semi-supervised classification, learning expressive variational distributions, and multimodal image-to-image translation from unpaired data.
\end{abstract}

\section{Introduction}
Deep generative models have achieved remarkable success in recent years. One of the most successful models is the generative adversarial network (GAN)~\citep{gan}, which employs a two player min-max game. The generative model, $G$, samples the noise vector $\mathbf{z} \sim p(\mathbf{z})$ and generates the sample $G(\mathbf{z})$. The discriminator, $D(\mathbf{x})$, is trained to identify whether a point $\mathbf{x}$ comes from the data distribution or the model distribution; and the generator is trained to maximally confuse the discriminator. The cost function of GAN is
\begin{align}
\underset{G}{\min}   \text{ } \underset{D}{\max} \text{ } \mathbb{E}_{\text{x} \sim p_{\text{data}}} [\log D(\mathbf{x})] + \mathbb{E}_{\mathbf{z} \sim p(\mathbf{z})} [\log (1 - D(G(\mathbf{\mathbf{z}}))].
\end{align}
GANs can be viewed as a general framework for learning implicit distributions~\citep{mohamed2016learning,ference}. Implicit distributions are probability distributions that are obtained by passing a noise vector through a deterministic function that is parametrized by a neural network. In the probabilistic machine learning problems, implicit distributions trained with the GAN framework can learn distributions that are more expressive than the tractable distributions trained with the maximum-likelihood framework.

Variational autoencoders (VAE)~\citep{vae,rezende} are another successful generative models that use neural networks to parametrize the posterior and the conditional likelihood distributions. Both networks are jointly trained to maximize a variational lower bound on the data log-likelihood. One of the limitations of VAEs is that they learn factorized distributions for both the posterior and the conditional likelihood distributions. In this paper, we propose the ``implicit autoencoder'' (IAE) that uses implicit distributions for learning more expressive posterior and conditional likelihood distributions. Learning a more expressive posterior will result in a tighter variational bound; and learning a more expressive conditional likelihood distribution will result in a high-level vs. low-level decomposition of information between the prior and the conditional likelihood. This enables the latent code to only capture the information that we care about such as the abstract or ``content'' information, while the remaining low-level or ``style'' information of data is separately captured by the noise vector of the implicit decoder.

Implicit distributions have been previously used in learning generative models in works such as adversarial autoencoders (AAE)~\citep{aae}, adversarial variational Bayes (AVB)~\citep{avb}, ALI~\citep{ali}, BiGAN~\citep{bigan} and other works such as~\citep{ference,him}. The high-level vs. low-level decomposition of information has also been studied in previous works such as PixelCNN autoencoders~\citep{pixelcnn}, PixelVAE~\citep{pixelvae}, variational lossy autoencoders~\citep{vlae}, PixelGAN autoencoders~\citep{pixelgan}, or other works such as~\citep{bowman,acn,alemifixing}.

In \mysec{iae}, we propose the IAE and then establish its connections with the related works. In \mysec{ciae}, we propose the cycle implicit autoencoder (CycleIAE), and show that it can learn multimodal cross-domain mappings from unpaired data. Finally, in \mysec{fiae}, we propose the flipped implicit autoencoder (FIAE) model, and show that it can learn expressive variational inference networks for GANs.

\section{Implicit Autoencoders}\label{sec:iae}

\begin{figure}[t]
\begin{center}
\includegraphics[scale=0.5]{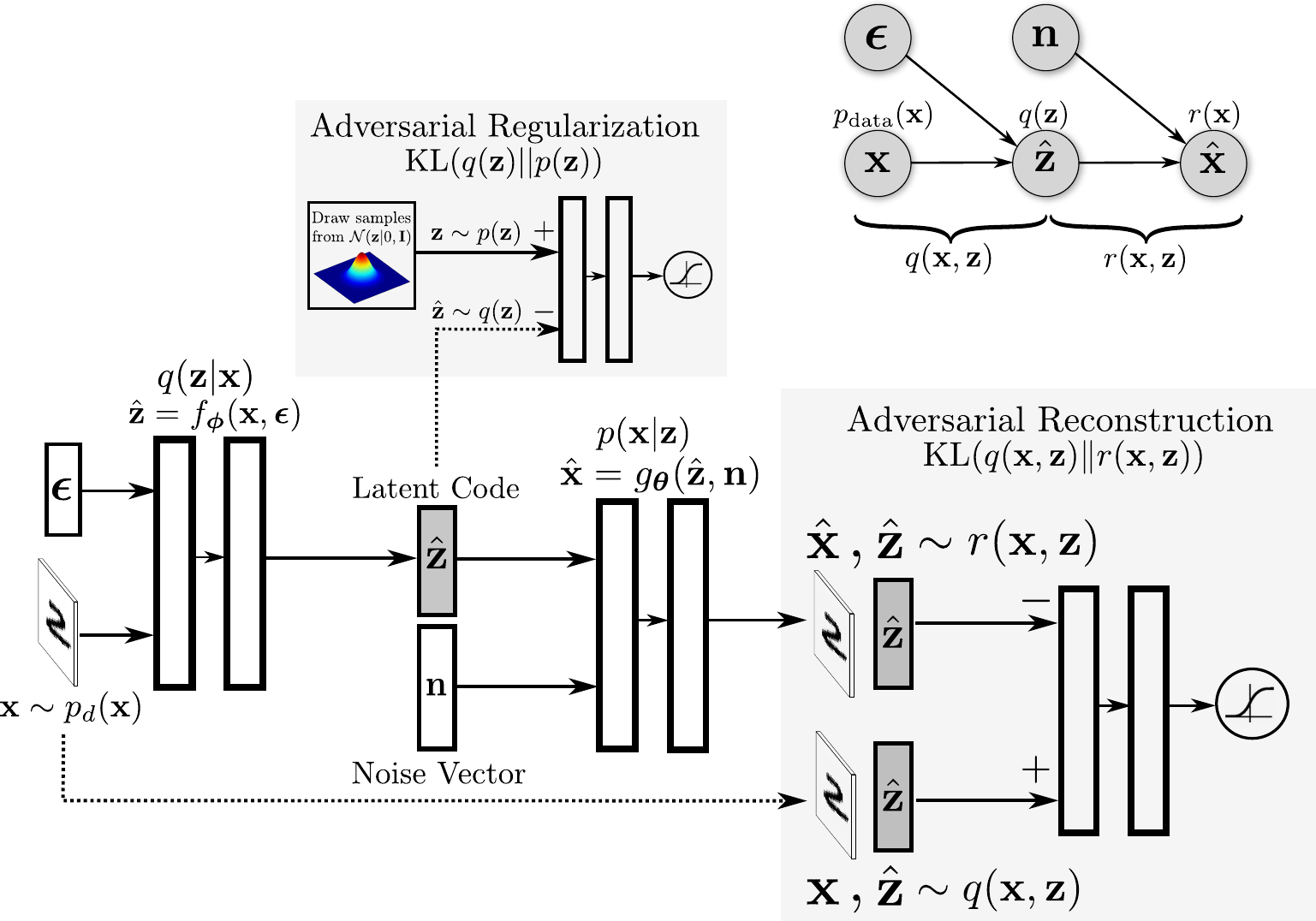}
\end{center}
\vspace{-.3cm}
\caption{\label{fig:iae}Architecture and graphical model of implicit autoencoders.}
\end{figure}

Let $\mathbf{x}$ be a datapoint that comes from the data distribution $p_\text{data}(\mathbf{x})$. The encoder of the implicit autoencoder (\myfig{iae}) defines an implicit variational posterior distribution $q(\mathbf{z}|\mathbf{x})$ with the function $\hat{\mathbf{z}} = f_{\bphi}(\mathbf{x},\bepsilon)$ that takes the input $\mathbf{x}$ along with the \emph{input noise vector} $\bepsilon$ and outputs the \emph{latent code} $\hat{\mathbf{z}}$. 
The decoder of the implicit autoencoder defines an implicit conditional likelihood distribution $p(\mathbf{x}|\mathbf{z})$ with the function $\hat{\mathbf{x}} = g_{\btheta}(\hat{\mathbf{z}},\mathbf{n})$ that takes the latent code $\hat{\mathbf{z}}$ along with the \emph{latent noise vector} $\mathbf{n}$ and outputs a reconstruction of the image $\hat{\mathbf{x}}$.
Let $p(\mathbf{z})$ be a fixed prior distribution, $p(\mathbf{x}, \mathbf{z})=p(\mathbf{z})p(\mathbf{x}| \mathbf{z})$ be the joint model distribution, and $p(\mathbf{x})$ be the model distribution. The variational distribution $q(\mathbf{z}|\mathbf{x})$ induces the \emph{joint data distribution} $q(\mathbf{x}, \mathbf{z})$, the \emph{aggregated posterior distribution} $q(\mathbf{z})$, and the \emph{inverse posterior/encoder distribution} $q(\mathbf{x}|\mathbf{z})$ as follows:
\begin{align}\quad q(\mathbf{x}, \mathbf{z}) = q(\mathbf{z}|\mathbf{x}) p_\text{data}(\mathbf{x}) \qquad
\hat{\mathbf{z}}\sim q(\mathbf{z}) = \int_\mathbf{x} q(\mathbf{x}, \mathbf{z}) d\mathbf{x} \qquad
q(\mathbf{x}| \mathbf{z}) = {q(\mathbf{x}, \mathbf{z}) \over q(\mathbf{z})}
\end{align}
Maximum likelihood learning is equivalent to matching the model distribution $p(\mathbf{x})$ to the data distribution $p_\text{data}(\mathbf{x})$; and learning with variational inference is equivalent to matching the joint model distribution $p(\mathbf{x}, \mathbf{z})$ to the joint data distribution $q(\mathbf{x}, \mathbf{z})$.
The entropy of the data distribution $\mathcal{H}_{\text{data}}{(\mathbf{x})}$, the entropy of the latent code $\mathcal{H}{(\mathbf{z})}$, the mutual information $\mathcal{I}(\mathbf{x};\mathbf{z})$, and the conditional entropies $\mathcal{H}{(\mathbf{x}|\mathbf{z})}$ and $\mathcal{H}{(\mathbf{z}|\mathbf{x})}$ are all defined under the joint data distribution $q(\mathbf{x}, \mathbf{z})$ and its marginals $p_\text{data}(\mathbf{x})$ and $q(\mathbf{z})$.
Using the aggregated posterior distribution $q(\mathbf{z})$, we can define the \emph{joint reconstruction distribution} $r(\mathbf{x}, \mathbf{z})$ and the \emph{aggregated reconstruction distribution} $r(\mathbf{x})$ as follows:
\begin{align}\quad r(\mathbf{x}, \mathbf{z}) = q(\mathbf{z})p(\mathbf{x}|\mathbf{z}) \qquad \hat{\mathbf{x}}\sim r(\mathbf{x}) = \int_\mathbf{z} r(\mathbf{x}, \mathbf{z}) d\mathbf{z}
\end{align}
Note that in general we have $r(\mathbf{x}, \mathbf{z}) \neq q(\mathbf{x}, \mathbf{z}) \neq p(\mathbf{x}, \mathbf{z})$, $q(\mathbf{z}) \neq p(\mathbf{z})$, and $r(\mathbf{x}) \neq p_\text{data}(\mathbf{x}) \neq p(\mathbf{x})$.

We now use different forms of the aggregated evidence lower bound (ELBO) to describe the IAE and establish its connections with VAEs and AAEs.
\begin{align}
\mathbb{E}_{\mathbf{x} \sim p_\text{data}(\mathbf{x})}[\log p(\mathbf{x})]
&\geq
-\underbrace{\mathbb{E}_{\mathbf{x} \sim p_\text{data}(\mathbf{x})}\Big[\mathbb{E}_{q(\mathbf{z}|\mathbf{x})} [-\log p(\mathbf{x}|\mathbf{z})]\Big]}_\text{VAE Reconstruction}
-\underbrace{\mathbb{E}_{\mathbf{x} \sim p_\text{data}(\mathbf{x})}\Big[\text{KL}( q(\mathbf{z}|\mathbf{x})\|p(\mathbf{z}))\Big]}_\text{VAE Regularization}\label{eq:vae}\\
&=
-\underbrace{\mathbb{E}_{\mathbf{x} \sim p_\text{data}(\mathbf{x})}\Big[\mathbb{E}_{q(\mathbf{z}|\mathbf{x})} [-\log p(\mathbf{x}|\mathbf{z})]\Big]}_\text{AAE Reconstruction} 
- \underbrace{\text{KL}( q(\mathbf{z})\|p(\mathbf{z}))}_\text{AAE Regularization}
- \underbrace{\mathcal{I}(\mathbf{z};\mathbf{x})}_\text{Mutual Info.} \label{eq:aae}\\
&=
-\underbrace{\mathbb{E}_{\mathbf{z} \sim q(\mathbf{z})}\Big[\text{KL}( q(\mathbf{x}|\mathbf{z})\|p(\mathbf{x}|\mathbf{z}))\Big]}_\text{IAE Reconstruction}
-\underbrace{\text{KL}( q(\mathbf{z})\|p(\mathbf{z}))}_\text{IAE Regularization}
-\underbrace{\mathcal{H}_{\text{data}}{(\mathbf{x})}}_\text{Entropy of Data} \label{eq:iae1} \\
&=
-\underbrace{\text{KL}(q(\mathbf{x},\mathbf{z})\|r(\mathbf{x},\mathbf{z}))}_\text{IAE Reconstruction}
-\underbrace{\text{KL}( q(\mathbf{z})\|p(\mathbf{z}))}_\text{IAE Regularization}
-\underbrace{\mathcal{H}_{\text{data}}{(\mathbf{x})}}_\text{Entropy of Data} \label{eq:iae2}
\end{align}
See \myappendix{proof} for the proof.
The standard formulation of the VAE (\myeq{vae}) only enables us to learn factorized posterior and conditional likelihood distributions. The AAE~\citep{aae} (\myeq{aae}) and the AVB~\citep{avb} enable us to learn implicit posterior distributions, but their conditional likelihood distribution is still a factorized distribution. However, the IAE enables us to learn implicit distributions for both the posterior and the conditional likelihood distributions. Similar to VAEs and AAEs, the IAE (\myeq{iae1}) has a reconstruction cost function and a regularization cost function, but trains each of them with a GAN. The IAE reconstruction cost is $\mathbb{E}_{\mathbf{z} \sim q(\mathbf{z})}\Big[\text{KL}( q(\mathbf{x}|\mathbf{z})\|p(\mathbf{x}|\mathbf{z}))\Big]$.
The standard VAE uses a factorized decoder, which has a very limited stochasticity, and thus achieves almost \emph{deterministic reconstructions} of the input. The IAE, however, uses a powerful implicit decoder to perform \emph{stochastic reconstructions}, by learning to match the expressive decoder distribution $p(\mathbf{x}| \mathbf{z})$ to the inverse encoder distribution $q(\mathbf{x}| \mathbf{z})$.
\myeq{aae_iae} contrasts the reconstruction cost of standard autoencoders that is used in VAEs/AAEs, with the reconstruction cost of IAEs.
\begin{align}
\underbrace{\mathbb{E}_{\mathbf{x} \sim p_\text{data}(\mathbf{x})}\Big[\mathbb{E}_{q(\mathbf{z}|\mathbf{x})} [-\log p(\mathbf{x}|\mathbf{z})]\Big]}_\text{AE Reconstruction} 
=
\underbrace{\mathbb{E}_{\mathbf{z} \sim q(\mathbf{z})}\Big[\text{KL}( q(\mathbf{x}|\mathbf{z})\|p(\mathbf{x}|\mathbf{z}))\Big]}_\text{IAE Reconstruction}
+\underbrace{\mathcal{H}(\mathbf{x}|\mathbf{z})}_\text{Cond. Entropy} \label{eq:aae_iae}
\end{align}
We can see from \myeq{aae_iae} that similar to IAEs, the reconstruction cost of the autoencoder encourages matching the decoder distribution to the inverse encoder distribution. But in autoencoders, the cost function also encourages minimizing the conditional entropy $\mathcal{H}(\mathbf{x}|\mathbf{z})$, or maximizing the mutual information $\mathcal{I}(\mathbf{x},\mathbf{z})$. Maximizing the mutual information in autoencoders enforces the latent code to capture both the high-level and low-level information. In contrast, in IAEs, the reconstruction cost does not penalize the encoder for losing the low-level information, as long as the decoder can invert the encoder distribution.
In order to minimize the reconstruction cost function of the IAE, we re-write it in the form of a distribution matching cost function between the joint data distribution and the joint reconstruction distribution $\text{KL}(q(\mathbf{x},\mathbf{z})\|r(\mathbf{x},\mathbf{z}))$ (\myeq{iae2}). This KL divergence is approximately minimized with the \emph{reconstruction GAN}.
The IAE has also a regularization cost function $\text{KL}( q(\mathbf{z})\|p(\mathbf{z}))$ that matches the aggregated posterior distribution with a fixed prior distribution. This is the same regularization cost function used in AAEs (\myeq{aae}), and is approximately minimized with the \emph{regularization GAN}. Note that the last term in \myeq{iae2} is the entropy of the data distribution that is fixed.

\parhead{Training Process.} We now describe the training process. We pass a given point $\mathbf{x} \sim p_\text{data}(\mathbf{x})$ through the encoder and the decoder to obtain $\hat{\mathbf{z}}\sim q(\mathbf{z})$ and $\hat{\mathbf{x}}\sim r(\mathbf{x})$. We now train the discriminator of the reconstruction GAN to identify the positive example $(\mathbf{x},\hat{\mathbf{z}})$ from the negative example $(\hat{\mathbf{x}},\hat{\mathbf{z}})$. This discriminator now defines the reconstruction cost of the IAE. We try to confuse this discriminator by backpropagating through the negative example $(\hat{\mathbf{x}},\hat{\mathbf{z}})$, and updating the encoder and decoder weights.
We call this process \emph{adversarial reconstruction}. Similarly, we train the discriminator of the regularization GAN to identify the positive example ${\mathbf{z}}\sim p(\mathbf{z})$ from the negative example $\hat{\mathbf{z}} \sim q(\mathbf{z})$. This discriminator now defines the regularization cost function, which can provide us with a gradient to update only the encoder weights. We call this process \emph{adversarial regularization}. Optimizing the adversarial regularization and reconstruction cost functions encourages $p(\mathbf{x}|\mathbf{z}) = q(\mathbf{x}|\mathbf{z})$ and $p(\mathbf{z}) = q(\mathbf{z})$, which results in the model distribution capturing the data distribution $p(\mathbf{x}) = p_\text{data}(\mathbf{x})$.

We note that in this work, we use the original formulation of GANs~\citep{gan} to match the distributions. As a result, the gradient that we obtain from the adversarial training, only approximately follows the gradient of the KL divergences and the variational bound. However, as shown in~\citep{f-gan}, by using the $f$-GAN objective in the GAN formulation, we can optimize any $f$-divergence including the KL divergence.

\parhead{High-Level vs. Low-Level Decomposition of Information in IAEs.} In IAEs, the dimension of the latent code along with its prior distribution defines the capacity of the latent code, and the dimension of the latent noise vector along with its distribution defines the capacity of the implicit decoder. By adjusting these dimensions and distributions, we can have a full control over the decomposition of information between the latent code and the implicit decoder. In one extreme case, by removing the noise vector, we can have a fully deterministic autoencoder that captures all the information by its latent code. In the other extreme case, we can remove the latent code and have an unconditional implicit distribution that can capture the whole data distribution by itself. 

In IAEs, we can choose to only optimize the reconstruction cost or both the reconstruction and the regularization costs. In the following, we discuss four special cases of the IAE and establish connections with the related methods.

\parhead{1. Deterministic Decoder without Regularization Cost}
\phantomsection
\label{case1}

In this case, we remove the noise vectors from the IAE, which makes both $q(\mathbf{z}| \mathbf{x})$ and $p(\mathbf{x}| \mathbf{z})$ deterministic. We then only optimize the reconstruction cost $\mathbb{E}_{\mathbf{z} \sim q(\mathbf{z})}\Big[\text{KL}( q(\mathbf{x}|\mathbf{z})\|p(\mathbf{x}|\mathbf{z}))\Big]$. As a result, similar to the standard autoencoder, the deterministic decoder $p(\mathbf{x}| \mathbf{z})$ learns to match to the inverse deterministic encoder $q(\mathbf{x}| \mathbf{z})$, and thus the IAE learns to perform exact and deterministic reconstruction of the original image, while the latent code is learned in an unconstrained fashion. In other words, in standard autoencoders, the Euclidean cost \emph{explicitly} encourages $\hat{\mathbf{x}}$ to reconstruct $\mathbf{x}$, and in case of uncertainty, performs mode averaging by blurring the reconstructions; however, in IAEs, the adversarial reconstruction \emph{implicitly} encourages $\hat{\mathbf{x}}$ to reconstruct $\mathbf{x}$, and in case of uncertainty, captures this uncertainty by the latent noise vector (\hyperref[case3]{Case 3}), which results in sharp reconstructions.

\parhead{2. Deterministic Decoder with Regularization Cost}

In the previous case, the latent code was learned in an unconstrained fashion. We now keep the decoder deterministic and add the regularization term which matches the aggregated posterior distribution to a fixed prior distribution. In this case, the IAE reduces to the AAE with the difference that the IAE performs adversarial reconstruction rather than Euclidean reconstruction. This case of the IAE defines a valid generative model where the latent code captures all the information of the data distribution. In order to sample from this model, we first sample from the imposed prior $p(\mathbf{z})$ and then pass this sample through the deterministic decoder.

\parhead{3. Stochastic Decoder without Regularization Cost}
\phantomsection
\label{case3}

In this case of the IAE, we only optimize ${\text{KL}(q(\mathbf{x},\mathbf{z})\|r(\mathbf{x},\mathbf{z}))}$, while $p(\mathbf{x}| \mathbf{z})$ is a stochastic implicit distribution. Matching the joint distribution $q(\mathbf{x}, \mathbf{z})$ to $r(\mathbf{x}, \mathbf{z})$ ensures that their marginal distributions would also match; that is, the aggregated reconstruction distribution $r(\mathbf{x})$ matches the data distribution $p_\text{data}(\mathbf{x})$. This model by itself defines a valid generative model in which both the prior, which in this case is $q(\mathbf{z})$, and the conditional likelihood $p(\mathbf{x}| \mathbf{z})$ are learned at the same time. In order to sample from this generative model, we initially sample from $q(\mathbf{z})$ by first sampling a point $\mathbf{x} \sim p_\text{data}(\mathbf{x})$ and then passing it through the encoder to obtain the latent code $\hat{\mathbf{z}} \sim q(\mathbf{z})$. Then we sample from the implicit decoder distribution conditioned on $\hat{\mathbf{z}}$ to obtain the stochastic reconstruction $\hat{\mathbf{x}} \sim r(\mathbf{x})$. If the decoder is deterministic (\hyperref[case1]{Case 1}), the reconstruction $\hat{\mathbf{x}}$ would be the same as the original image $\mathbf{x}$. But if the decoder is stochastic, the latent code only captures the abstract and high-level information of the image, and the stochastic reconstruction $\hat{\mathbf{x}}$ only shares this high-level information with the original $\mathbf{x}$. This case of the IAE is related to the PixelCNN autoencoder~\citep{pixelcnn}, where the decoder is parametrized by an autoregressive neural network which can learn expressive distributions, while the latent code is learned in an unconstrained fashion.

\parhead{4. Stochastic Decoder with Regularization Cost} 

In the previous case, we showed that even without the regularization term, $r(\mathbf{x})$ will capture the data distribution. But the main drawback of the previous case is that its prior $q(\mathbf{z})$ is not a parametric distribution that can be easily sampled from. One way to fix this problem is to fit a parametric prior $p(\mathbf{z})$ to $q(\mathbf{z})$ once the training is complete, and then use $p(\mathbf{z})$ to sample from the model. However, a better solution would be to consider a fixed and pre-defined prior $p(\mathbf{z})$, and impose it on $q(\mathbf{z})$ during the training process. Indeed, this is the regularization term that the ELBO suggests in \myeq{iae2}. By adding the adversarial regularization cost function to match $q(\mathbf{z})$ to $p(\mathbf{z})$, we ensure that $r(\mathbf{x}) = p_\text{data}(\mathbf{x}) = p(\mathbf{x})$. Now sampling from this model only requires first sampling from the pre-defined prior $\mathbf{z} \sim p(\mathbf{z})$, and then sampling from the conditional implicit distribution to obtain $\hat{\mathbf{x}} \sim r(\mathbf{x})$. In this case, the information of data distribution is captured by both the fixed prior and the learned conditional likelihood distribution. Similar to the previous case, the latent code captures the high-level and abstract information, while the remaining low-level information is captured by the implicit decoder. We will empirically show this decomposition of information on different datasets in \mysec{high-level} and \mysec{discrete}. This decomposition of information has also been studied in other works such as PixelVAE~\citep{pixelvae}, variational lossy autoencoders~\citep{vlae}, PixelGAN autoencoders~\citep{pixelgan} and variational Seq2Seq autoencoders~\citep{bowman}. However, the main drawback of these methods is that they all use autoregressive decoders which are not parallelizable, and are much more computationally expensive to scale up than the implicit decoders. Another advantage of implicit decoders to autoregressive decoders is that in implicit decoders, the low-level statistics is captured by the noise vector representation; but in autoregressive decoders, there is no vector representation for the low-level statistics.

\parhead{Connections with ALI and BiGAN.}
In ALI~\citep{ali} and BiGAN~\citep{bigan} models, there are two separate networks that define the joint data distribution $q(\mathbf{x}, \mathbf{z})$ and the joint model distribution $p(\mathbf{x}, \mathbf{z})$. The parameters of these networks are trained using the gradient that comes from a single GAN that tries to match these two distributions. However, in the IAE, similar to VAEs or AAEs, the encoder and decoder are stacked on top of each other and trained jointly. So the gradient that the encoder receives comes through the decoder and the conditioning vector. In other words, in the ALI model, the input to the conditional likelihood is the samples of the prior distribution, whereas in the IAE, the input to the conditional likelihood is the samples of the variational posterior distribution, while the prior distribution is separately imposed on the aggregated posterior distribution by the regularization GAN. This makes the training dynamic of IAEs similar to that of autoencoders, which encourages better reconstructions. 

\subsection{Experiments of Implicit Autoencoders}
\subsubsection{High-Level vs. Low-Level Decomposition of Information}\label{sec:high-level}

\begin{figure}[t]
\centering
\subfigure[\hspace{-.2cm}]{
\includegraphics[scale=.13]{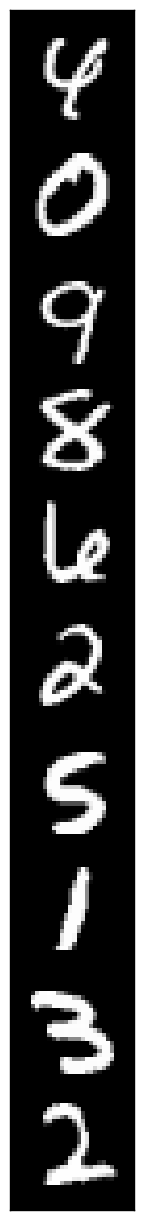}}
\subfigure[\hspace{-.2cm}]{
\includegraphics[scale=.13]{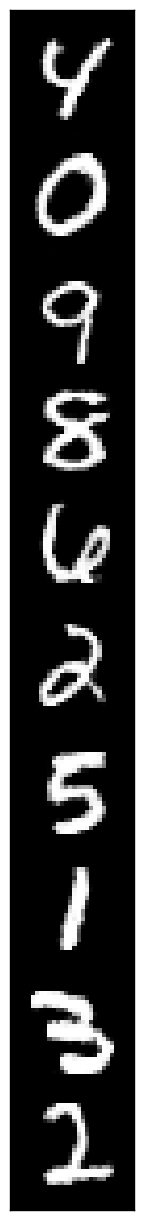}}
\subfigure[\hspace{-.2cm}]{
\includegraphics[scale=.2]{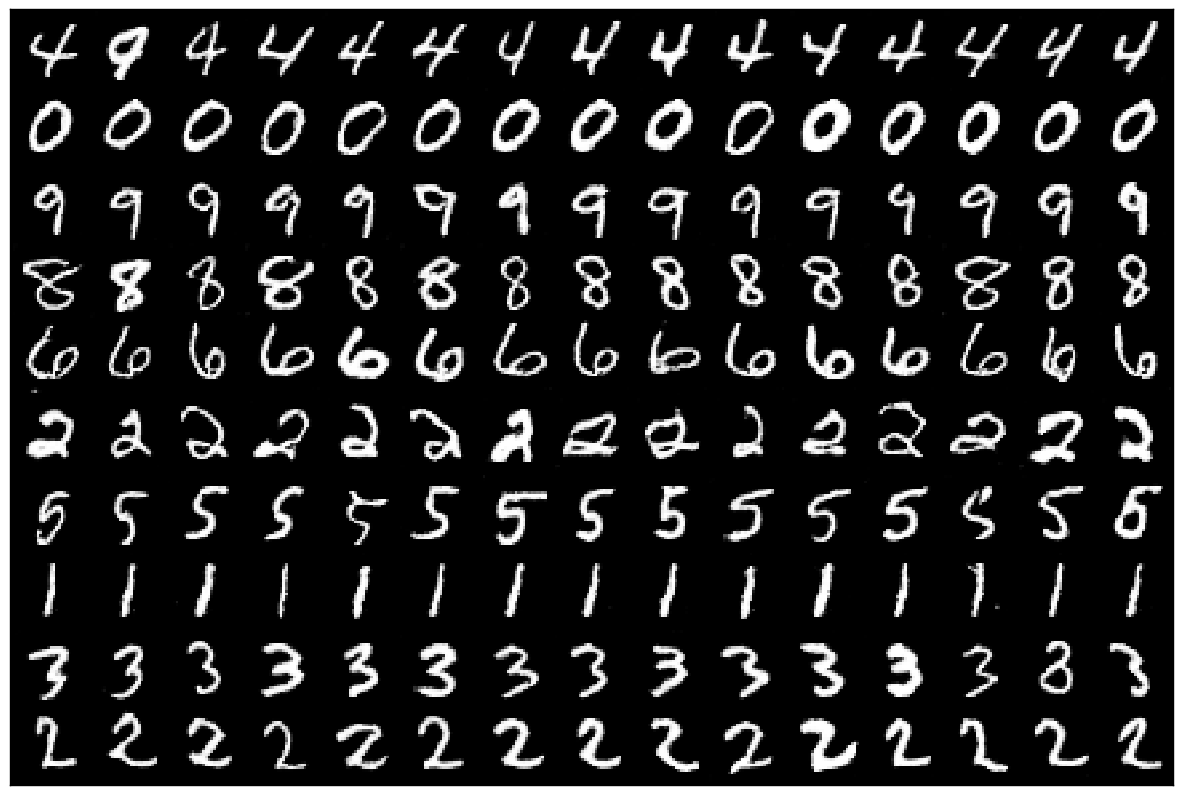}}
\subfigure[\hspace{-.2cm}]{
\includegraphics[scale=.2]{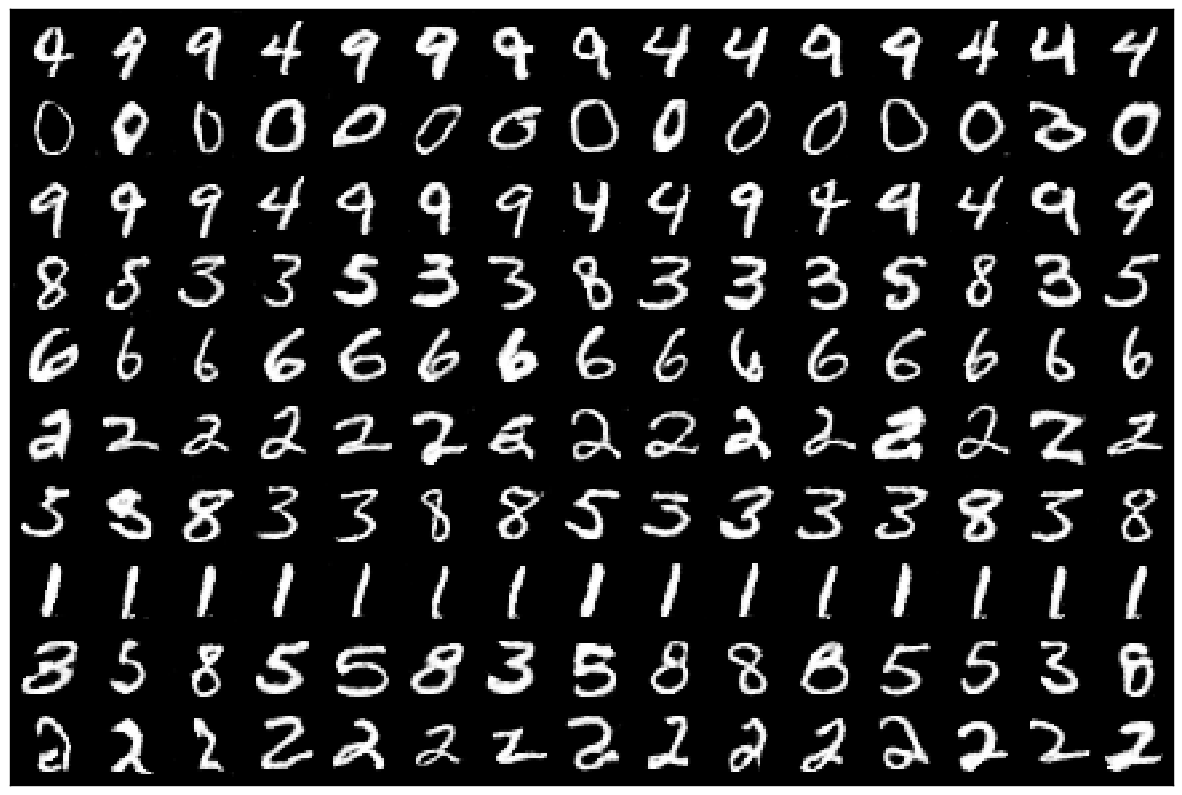}}
\vspace{-.3cm}
\caption{\label{fig:iae_mnist}MNIST dataset. (a) Original images. (b) Deterministic reconstructions with 20D latent code. (c) Stochastic reconstructions with 10D latent code and 100D latent noise vector. (d) Stochastic reconstructions with 5D latent code and 100D latent noise vector.}
\end{figure}

\begin{figure}[t]
\begin{minipage}{\linewidth}
    \centering
    \begin{minipage}{0.49\linewidth}
        \begin{figure}[H]
          \subfigure[\hspace{-.2cm}]{
          \includegraphics[scale=.169]{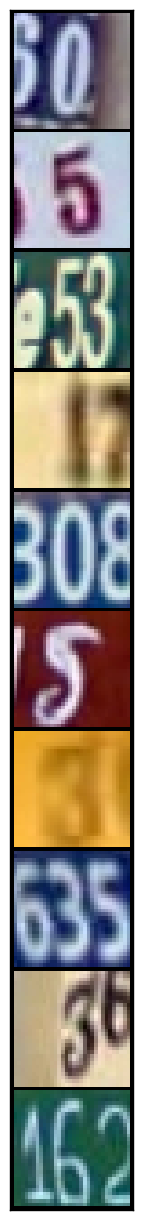}}
          \hspace{-.2cm}
          \subfigure[\hspace{-.2cm}]{
          \includegraphics[scale=.169]{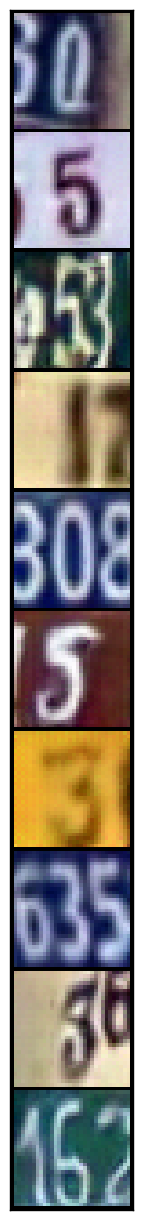}} 
          \hspace{-.2cm}
          \subfigure[\hspace{-.2cm}]{
          \includegraphics[scale=.1745]{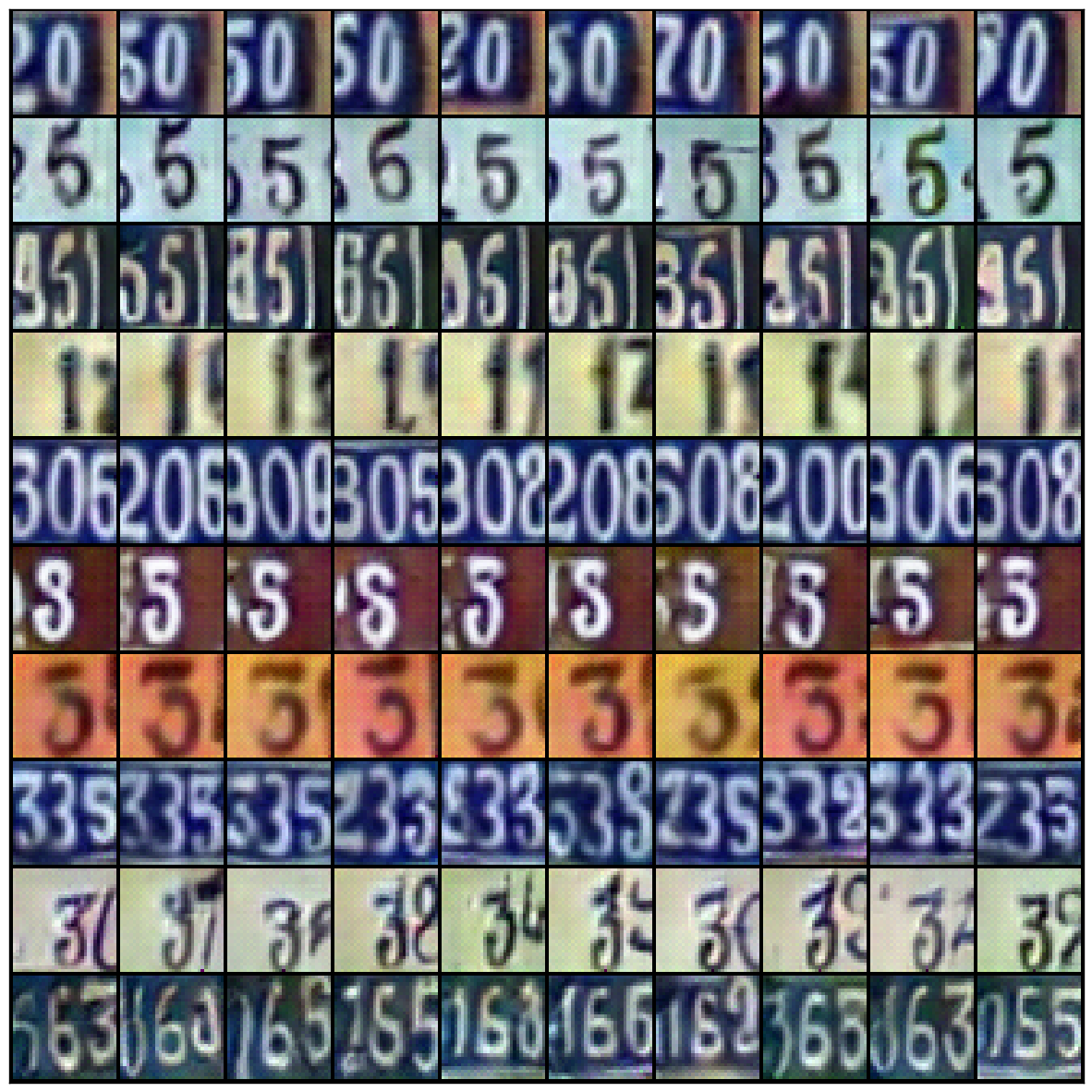}}
          \vspace{-.3cm}
          \caption{\label{fig:iae_svhn}SVHN dataset. (a) Original images. (b) Deterministic reconstructions with 150D latent code. (c) Stochastic reconstructions with 75D latent code and 1000D latent noise vector.}
        \end{figure}
    \end{minipage}
    \hspace{.1cm}
    \begin{minipage}{0.49\linewidth}
        \begin{figure}[H]
          \subfigure[\hspace{-.2cm}]{
          \includegraphics[scale=.169]{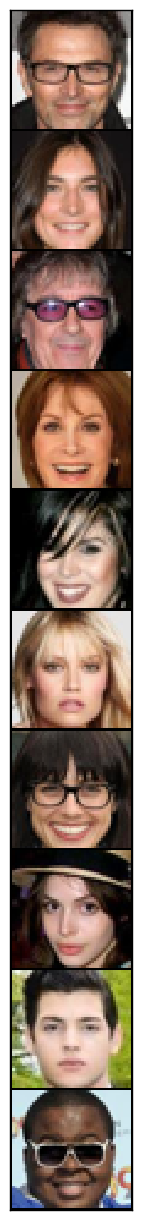}}
          \hspace{-.2cm}          
          \subfigure[\hspace{-.2cm}]{
          \includegraphics[scale=.169]{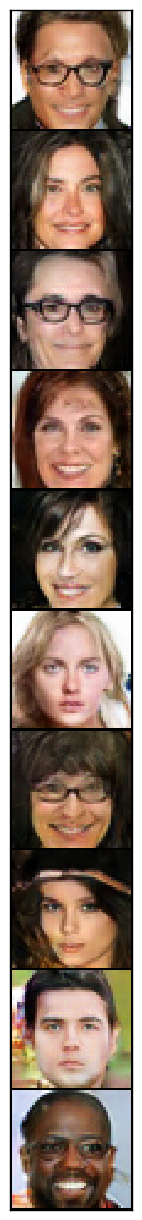}}
          \hspace{-.2cm}          
          \subfigure[\hspace{-.2cm}]{
          \includegraphics[scale=.1745]{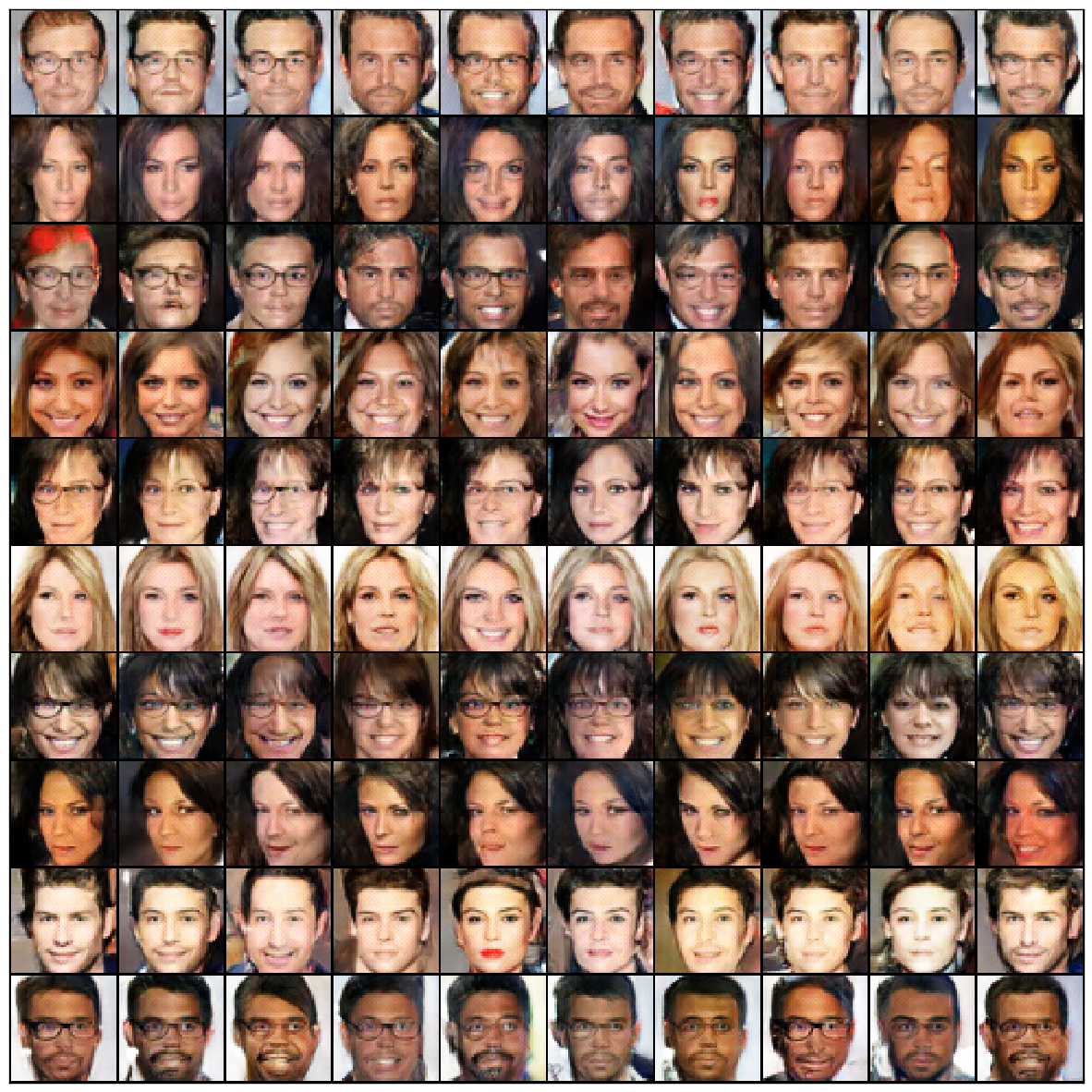}}
          \vspace{-.3cm}
          \caption{\label{fig:iae_celeb}CelebA dataset. (a) Original images. (b) Deterministic reconstructions with 150D latent code. (c) Stochastic reconstructions with 50D latent code and 1000D latent noise vector.}
          \end{figure}
    \end{minipage}
\end{minipage}
\end{figure}

In this section, we show that the IAE can learn a high-level vs. low-level decomposition of information between the latent code and the implicit decoder.
We use the Gaussian distribution for both the latent code and noise vectors, and show that by adjusting the dimensions of the latent code and the noise vector, we can have a full control over the decomposition of information.

\myfig{iae_mnist} shows the performance of the IAE on the MNIST dataset. By removing the noise vector and using only a latent code of size 20D (\myfigg{iae_mnist}{b}), the IAE becomes a deterministic autoencoder. In this case, the latent code of the IAE captures all the information of the data distribution and the IAE achieves almost perfect reconstructions. By decreasing the latent code size to 10D and using a 100D noise vector (\myfigg{iae_mnist}{c}), the latent code retains the high-level information of the digits such as the label information, while the noise vector captures small variations in the style of the digits. By using a smaller latent code of size 5D (\myfigg{iae_mnist}{d}), the encoder loses more low-level information and thus the latent code captures more abstract information. For example, we can see from \myfigg{iae_mnist}{d} that the encoder maps visually similar digits such as $\{3,5,8\}$ or $\{4,9\}$ to the same latent code, while the implicit decoder learns to invert this mapping and generate stochastic reconstructions that share the same high-level information with the original images. Note that if we completely remove the latent code, the noise vector captures all the information, similar to the standard unconditional GAN.

\myfig{iae_svhn} shows the performance of the IAE on the SVHN dataset. When using a 150D latent code with no noise vector (\myfigg{iae_svhn}{b}), similar to the standard autoencoder, the IAE captures all the information by its latent code and can achieve almost perfect reconstructions. However, when using a 75D latent code along with a 1000D noise vector (\myfigg{iae_svhn}{c}), the latent code of the IAE only captures the middle digit information as the high-level information, and loses the left and right digit information. At the same time, the implicit decoder learns to invert the encoder distribution by keeping the middle digit and generating synthetic left and right SVHN digits with the same style of the middle digit.

\myfig{iae_celeb} shows the performance of the IAE on the CelebA dataset. When using a 150D latent code with no noise vector (\myfigg{iae_celeb}{b}), the IAE achieves almost perfect reconstructions. But when using a 50D latent code along with a 1000D noise vector (\myfigg{iae_celeb}{c}), the latent code of the IAE only retains the high-level information of the face such as the general shape of the face, while the noise vector captures the low-level attributes of the face such as eyeglasses, mustache or smile.

\subsubsection{Clustering and Semi-Supervised Learning}\label{sec:discrete}

In IAEs, by using a categorical latent code along with a Gaussian noise vector, we can disentangle the discrete and continuous factors of variation, and perform clustering and semi-supervised learning.

\parhead{Clustering.} In order to perform clustering with IAEs, we change the architecture of \myfig{iae} by using a softmax function in the last layer of the encoder, as a continuous relaxation of the categorical latent code. The dimension of the categorical code is the number of categories that we wish the data to be clustered into. The regularization GAN is trained directly on the continuous output probabilities of the softmax simplex, and imposes the categorical distribution on the aggregated posterior distribution. This adversarial regularization imposes two constraints on the encoder output. The first constraint is that the encoder has to make confident decisions about the cluster assignments. The second constraint is that the encoder must distribute the points evenly across the clusters. As a result, the latent code only captures the discrete underlying factors of variation such as class labels, while the rest of the structure of the image is separately captured by the Gaussian noise vector of the implicit decoder.

\begin{figure}[t]
\centering
\subfigure[Original data]{
\includegraphics[scale=.1]{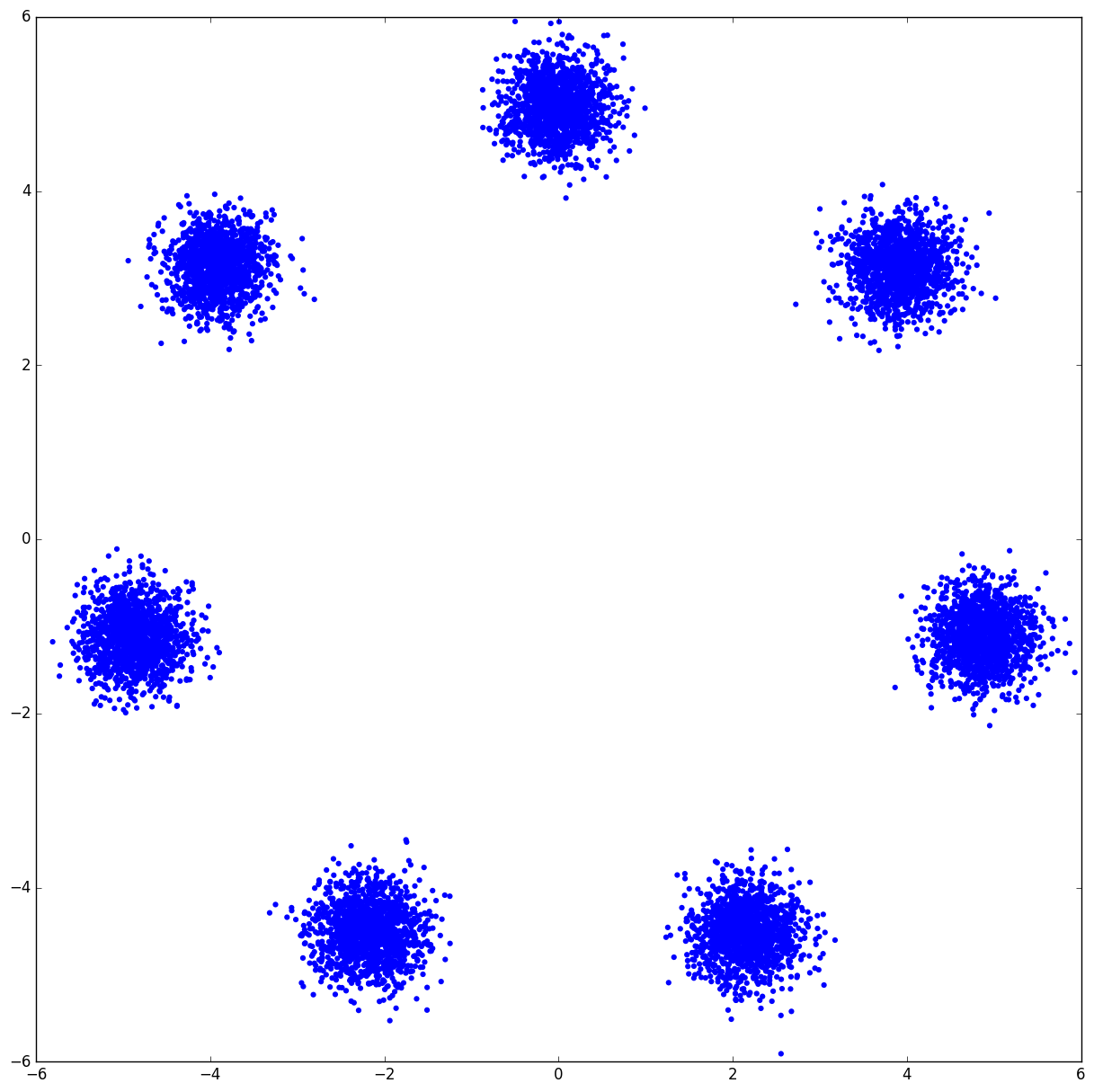}}
\hspace{1cm}
\subfigure[GAN]{
\includegraphics[scale=.1]{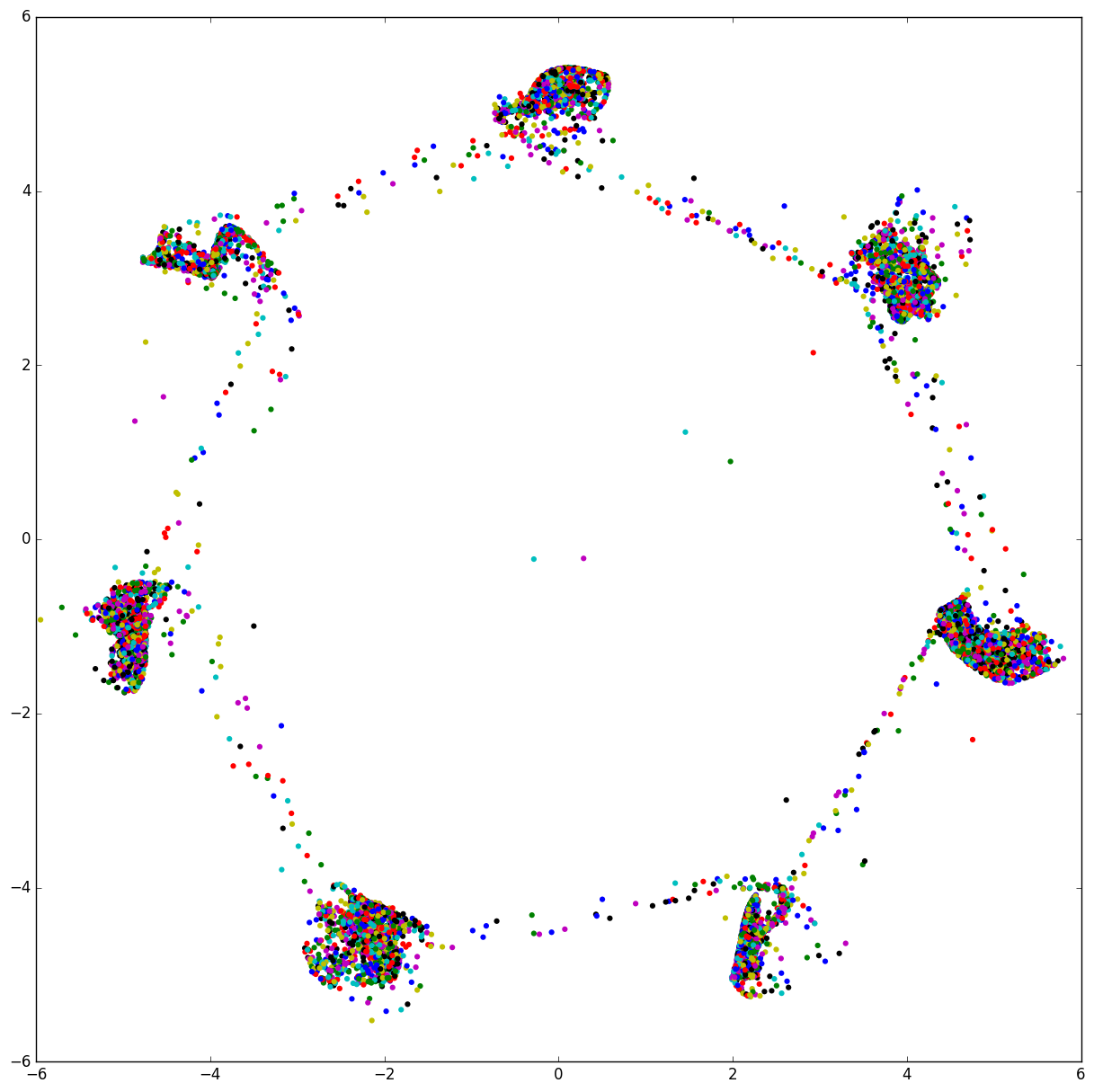}}
\hspace{1cm}
\subfigure[IAE]{
\includegraphics[scale=.1]{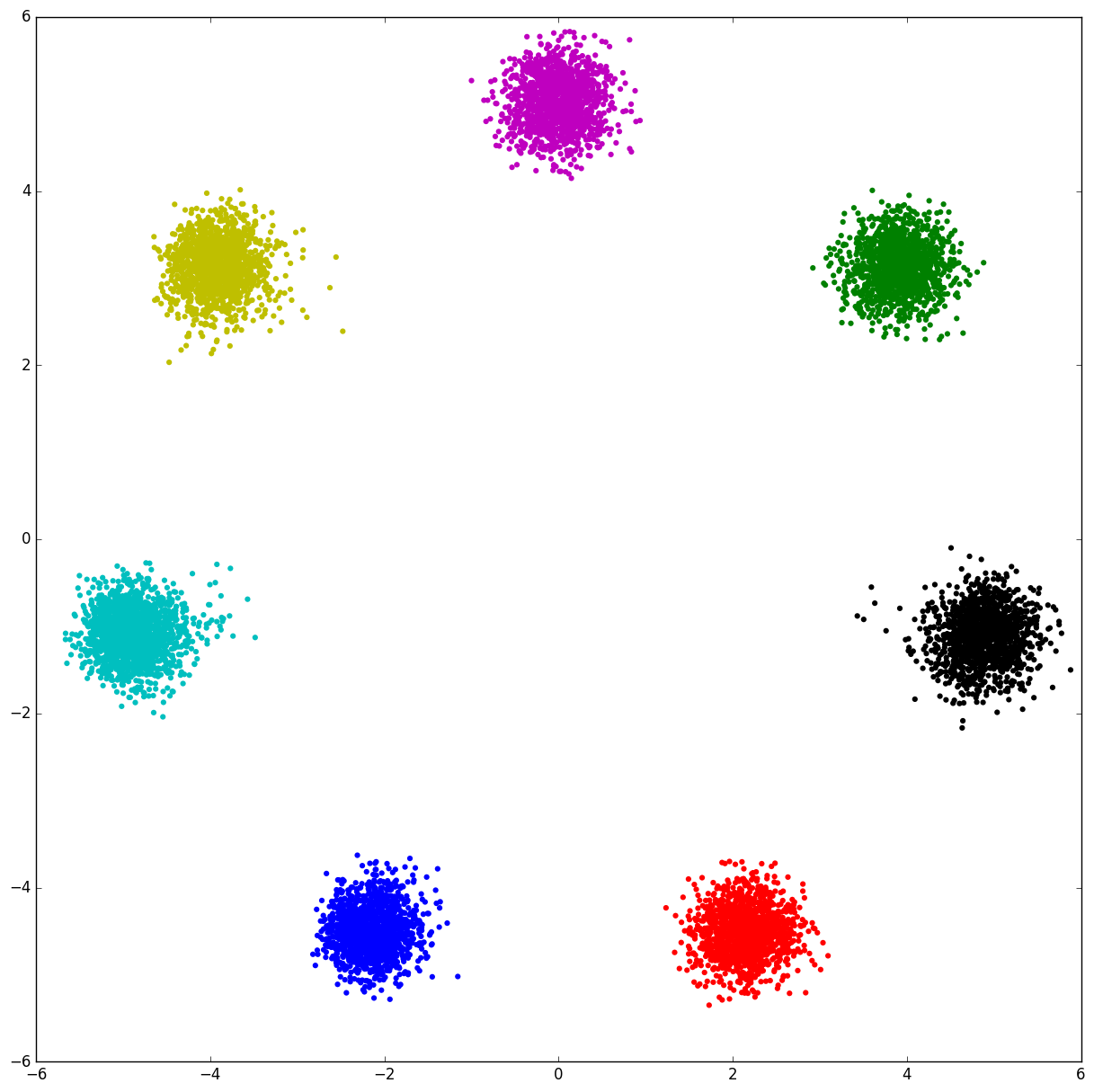}}
\vspace{-.3cm}
\caption{\label{fig:mog}Learning the mixture of Gaussian distribution by the standard GAN and the IAE.}
\end{figure}

\myfig{mog} shows the samples of the standard GAN and the IAE trained on the mixture of Gaussian data. \myfigg{mog}{b} shows the samples of the GAN, which takes a 7D categorical and a 10D Gaussian noise vectors as the input.
Each sample is colored based on the one-hot noise vector that it was generated from. We can see that the GAN has failed to associate the categorical noise vector to different mixture components, and generate the whole data solely by using its Gaussian noise vector. Ignoring the categorical noise forces the GAN to do a continuous interpolation between different mixture components, which results in reducing the quality of samples.
\myfigg{mog}{c} shows the samples of the IAE whose implicit decoder architecture is the same as the GAN. The IAE has a 7D categorical latent code (inferred by the encoder) and a 10D Gaussian noise vector. In this case, the inference network of the IAE learns to cluster the data in an unsupervised fashion, while its generative path learns to condition on the inferred cluster labels and generate each mixture component using the stochasticity of the Gaussian noise vector. This example highlights the importance of using discrete latent variables for improving generative models. A related work is the InfoGAN~\citep{infogan}, which uses a reconstruction cost in the code space to prevent the GAN from ignoring the categorical noise vector. The relationship of InfoGANs with IAEs is discussed in details in \mysec{fiae}.

\begin{figure}[b]
\centering
\includegraphics[scale=.3]{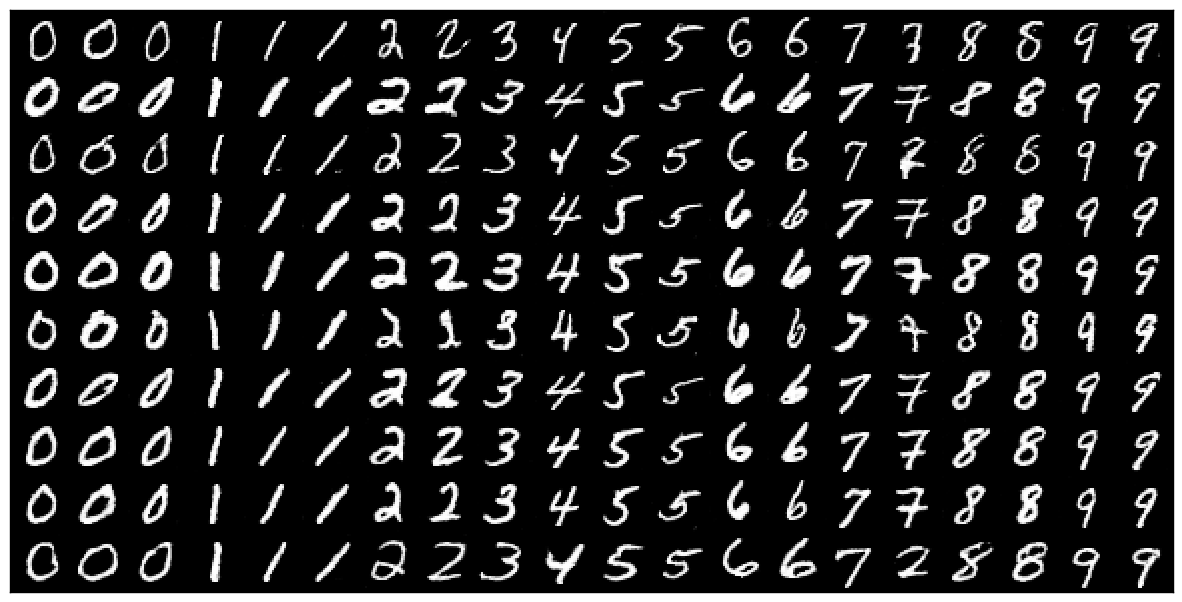}
\vspace{-.3cm}
\caption{\label{fig:mnist_cluster}Disentangling the content and style of the MNIST digits in an unsupervised fashion with implicit autoencoders. Each column shows samples of the model from one of the learned clusters. The style (latent noise vector) is drawn from a Gaussian distribution and held fixed across each row.}
\end{figure}

\myfig{mnist_cluster} shows the clustering performance of the IAE on the MNIST dataset. The IAE has a 30D categorical latent code and a 10D Gaussian noise vector. Each column corresponds to the conditional samples from one of the learned clusters (only 20 are shown). The noise vector is sampled from the Gaussian distribution and held fixed across each row. We can see that the discrete latent code of the network has learned discrete factors of variation such as the digit identities, while the writing style information is separately captured by the continuous Gaussian noise vector.
This network obtains about 5\% error rate in classifying digits in an unsupervised fashion, just by matching each cluster to a digit type.

\parhead{Semi-Supervised Learning.} The IAE can be used for semi-supervised classification. In order to incorporate the label information, we set the number of clusters to be the same as the number of class labels and additionally train the encoder weights on the labeled mini-batches to minimize the cross-entropy cost. On the MNIST dataset with 100 labels, the IAE achieves the error rate of $1.40\%$. In comparison, the AAE achieves $1.90\%$, and the Improved-GAN~\citep{improved-gan} achieves $0.93\%$. On the SVHN dataset with 1000 labels, the IAE achieves the error rate of $9.80\%$. In comparison, the AAE achieves $17.70\%$, and the Improved-GAN achieves $8.11\%$.

\section{Cycle Implicit Autoencoders}\label{sec:ciae}
The problem of image-to-image translation from unpaired data is an important problem in machine learning with many applications such as super-resolution, colorization, and attribute/style transfer.
Inspired by generative adversarial networks, recently several methods such as CycleGAN~\citep{cyclegan} have been proposed for learning cross-domain mappings. CycleGANs learn invertible cross-domain mappings by combining the GAN formulation with a cycle-consistency constraint that encourages the reconstruction of the input. Therefore, the CycleGAN model can be viewed as an adversarial autoencoder, where the network autoencodes the first domain using the cycle-consistency constraint, while at the same time, the empirical distribution of the second domain (instead of the prior) is imposed on the latent code of the autoencoder using the GAN framework. A major limitation of the CycleGAN/AAE is that they can only learn deterministic one-to-one mappings between domains, which is the result of the unimodality of the Euclidean or $L_1$ reconstruction error used in these networks. However, in many scenarios, the cross-domain mapping between two domains are inherently multi-modal or many-to-many, and thus can be better characterized using stochastic functions. For example in the edges$\leftrightarrow$shoes dataset~\citep{pix2pix}, an image in the edges domain can be potentially mapped to many images of the shoes with different styles. In this work, we showed that IAEs generalize adversarial autoencoders by replacing the Euclidean reconstruction cost with the adversarial reconstruction cost, enabling us to learn stochastic encoder/decoder functions. 
In this section, we use the similar idea to propose the CycleIAE, which replaces the $L_1$ cycle-consistency cost of the CycleGAN with an adversarial cycle-consistency cost, enabling us to learn stochastic image-to-image mappings.

\begin{figure}[t]
\begin{center}
\includegraphics[scale=0.5]{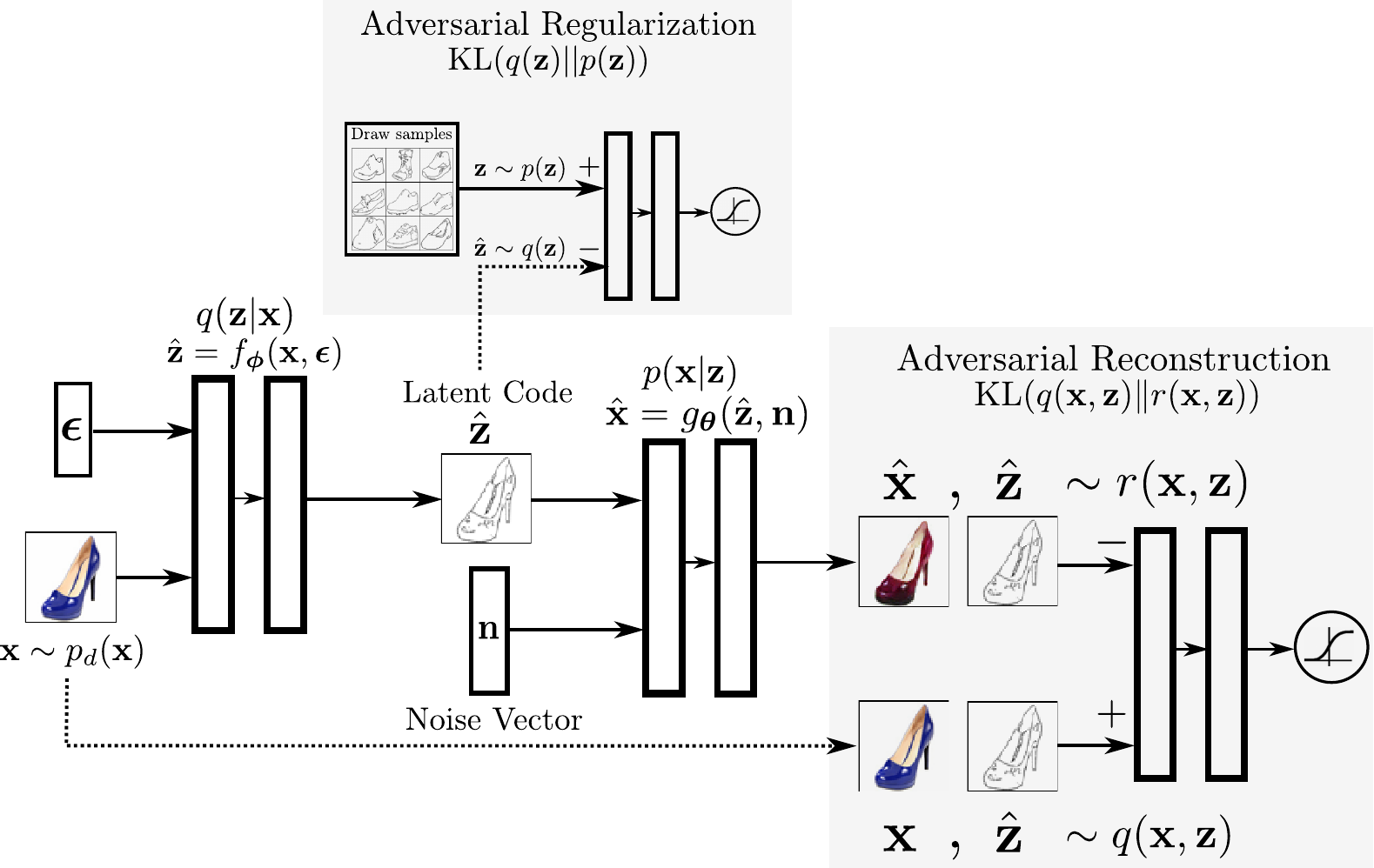}
\end{center}
\vspace{-.3cm}
\caption{\label{fig:cycleiae}Architecture of cycle implicit autoencoders.}
\end{figure}

\begin{figure}[t]
\begin{minipage}{\linewidth}
    \centering
    \begin{minipage}{0.49\linewidth}
        \begin{figure}[H]
          \subfigure[\hspace{-.2cm}]{
          \includegraphics[scale=.169]{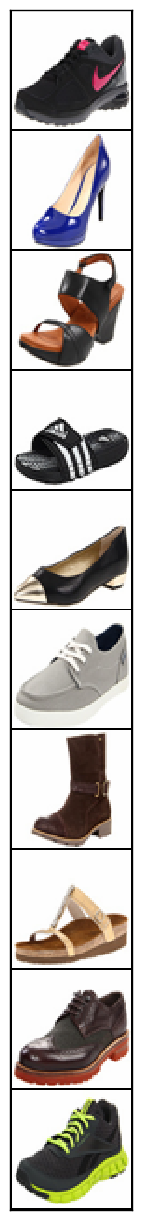}}
          \hspace{-.2cm}
          \subfigure[\hspace{-.2cm}]{
          \includegraphics[scale=.169]{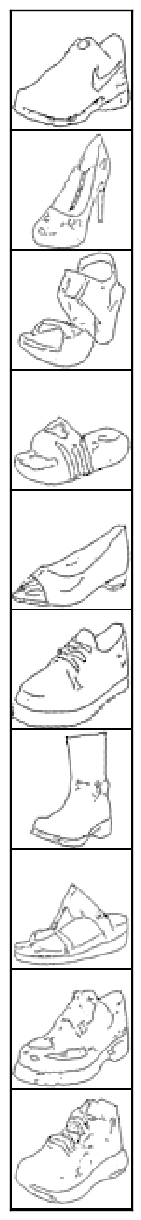}} 
          \hspace{-.2cm}
          \subfigure[\hspace{-.2cm}]{
          \includegraphics[scale=.1745]{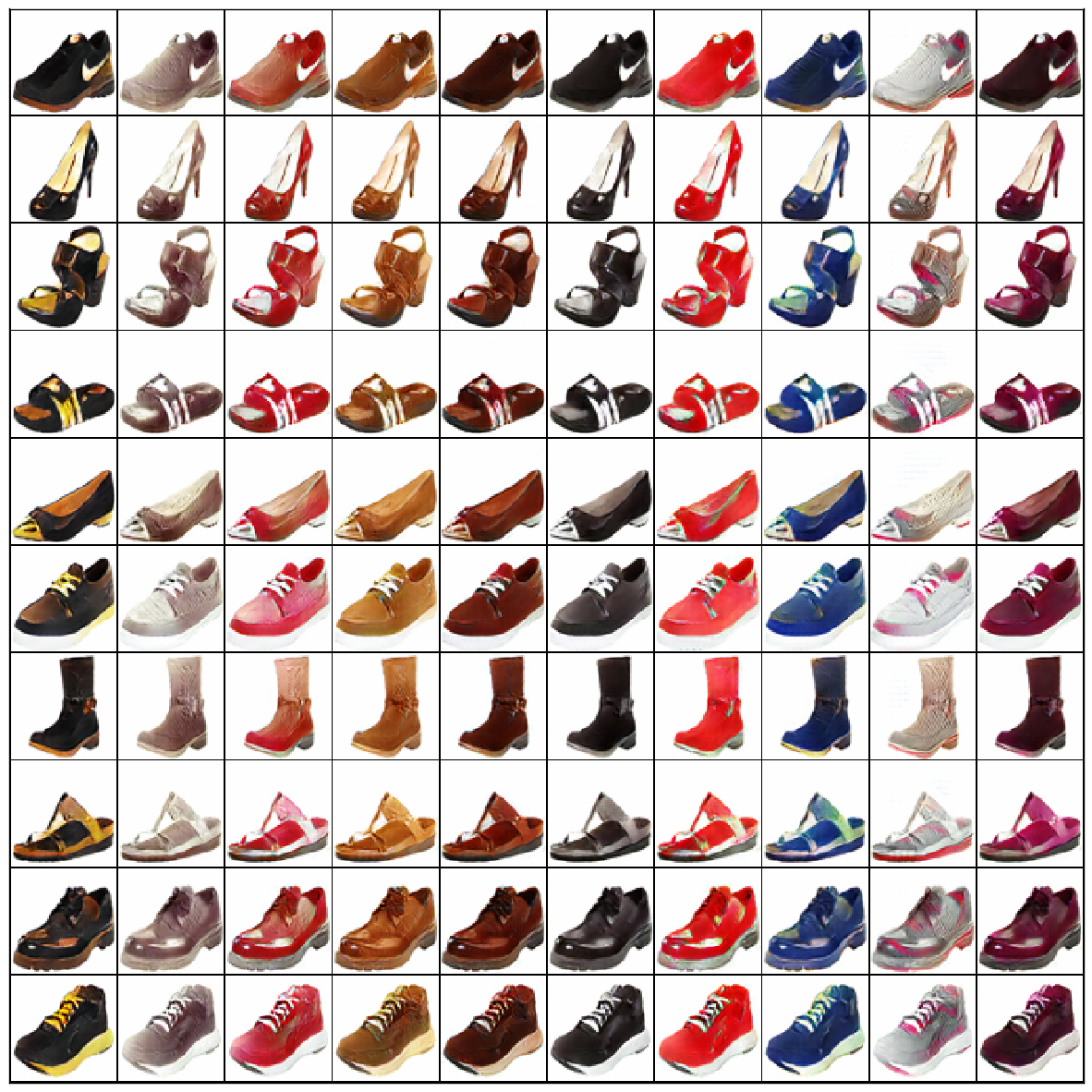}}
          \vspace{-.3cm}
  		  \caption{\label{fig:shoe}Training a CycleIAE that autoencodes shoe images while the distribution of edge images is imposed on the latent code. The encoder is deterministic and the decoder is stochastic. (a) Original shoe images as the input. (b) Inferred deterministic latent code. (c) Stochastic reconstructions with a noise vector of size 100. The latent code is held fixed across each row and the noise vector is drawn from a Gaussian distribution and held fixed across each column.}
        \end{figure}
    \end{minipage}
    \hspace{.1cm}
    \begin{minipage}{0.49\linewidth}
        \begin{figure}[H]
          \subfigure[\hspace{-.2cm}]{
          \includegraphics[scale=.169]{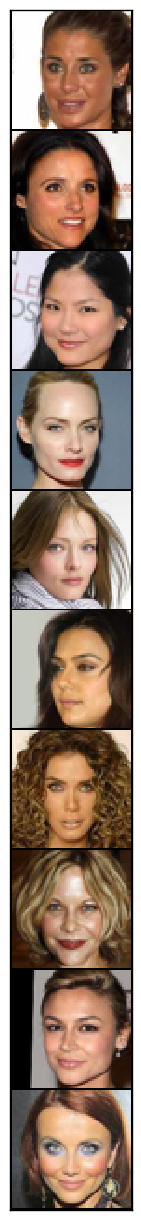}}
          \hspace{-.2cm}          
          \subfigure[\hspace{-.2cm}]{
          \includegraphics[scale=.1745]{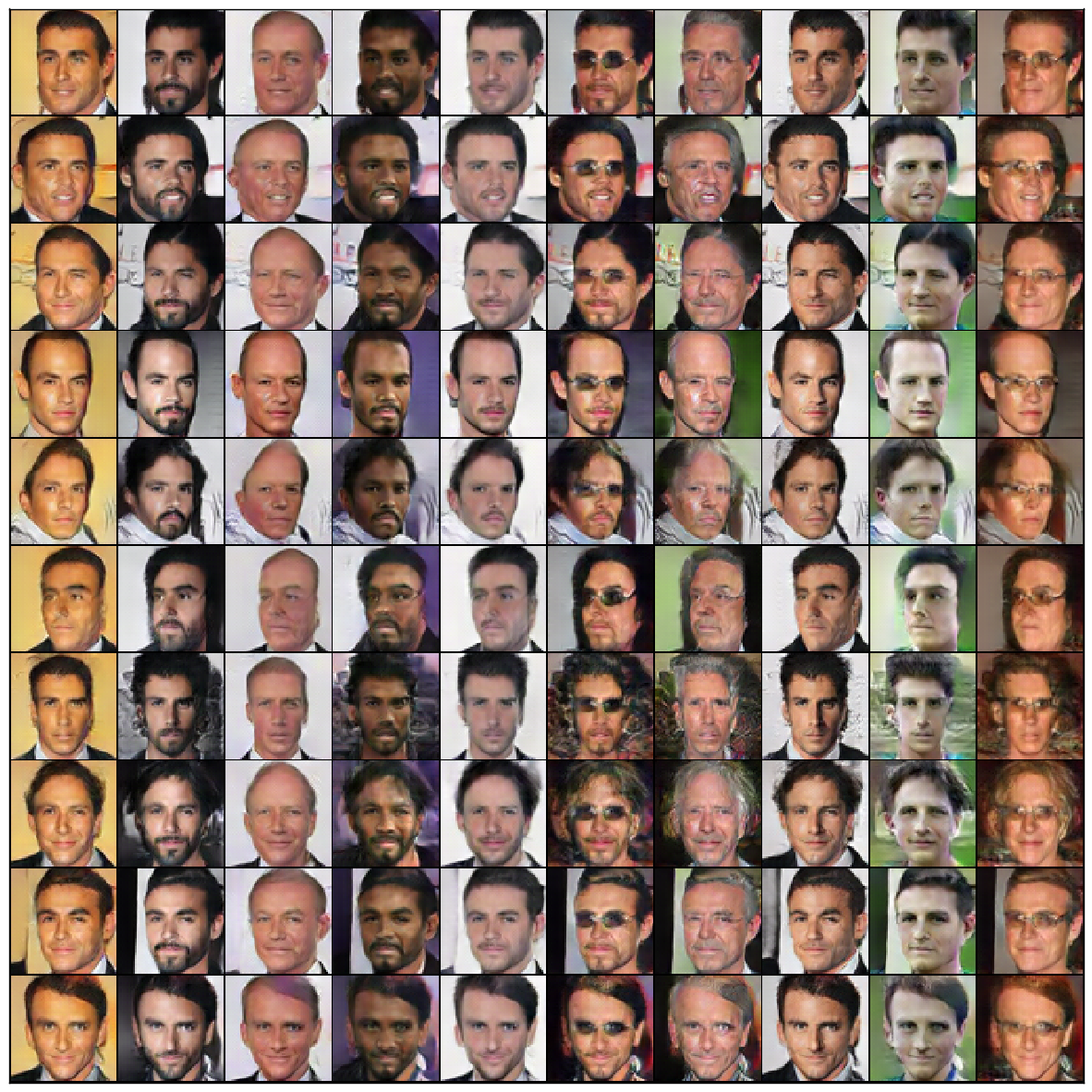}}
          \hspace{-.2cm}          
          \subfigure[\hspace{-.2cm}]{
          \includegraphics[scale=.169]{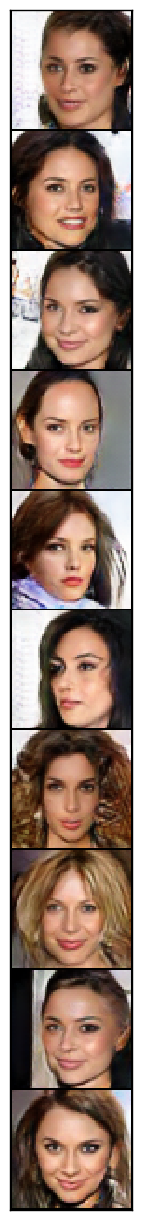}}
          \vspace{-.3cm}
  		  \caption{\label{fig:cycle-celeb}Training a CycleIAE that autoencodes female face images while the distribution of male face images is imposed on the latent code. The encoder is stochastic and the decoder is deterministic. (a) Original female face images as the input. (b) Samples of the stochastic latent code, where the noise vector is drawn from a Gaussian distribution of size 100 and held fixed across each column. (c) Deterministic reconstructions of the original images.}
          \end{figure}
    \end{minipage}
\end{minipage}
\end{figure}

The CycleIAE architecture is shown in \myfig{cycleiae}. In CycleIAEs, both the encoder and decoder are implicit distributions that are trained with the adversarial regularization and the adversarial reconstruction terms. The network autoencodes the distribution of the first domain using the adversarial reconstruction cost, while the distribution of the second domain is imposed on the latent code using the adversarial regularization cost. 

\myfig{shoe} illustrates the results of training the CycleIAE on the edges$\leftrightarrow$shoes dataset. In this network, we have used a deterministic encoder (no noise vector at the input) and a stochastic decoder (noise vector of size 100). The original shoe images are fed to the deterministic encoder, which learns to lose the style information of the shoe and infer the content or the edge image of the shoe in an unsupervised fashion (\myfigg{shoe}{b}). At the same time, the stochastic decoder learns to invert the encoder mapping by conditioning on the content information and generating stochastic reconstructions of the input image with different styles using the stochasticity of the noise vector (\myfigg{shoe}{c}). Note that if we use the CycleGAN's $L_1$ cycle-consistency cost instead of the joint distribution matching reconstruction cost, the network would learn to ignore the noise vector and perform deterministic reconstruction of the original image, due to the unimodality of the $L_1$ reconstruction error. This mode collapse behavior has been studied in details in~\citep{aug-cyclegan}.

Now we study the task of translating female faces to male faces in the CelebA dataset. \myfig{cycle-celeb} illustrates the results of training the CycleIAE, where the network learns to autoencode the female faces while the distribution of male faces is imposed on the latent code. In this experiment, we choose to have a stochastic encoder (noise vector of size 100) and a deterministic decoder (no noise vector at the input of the decoder). \myfigg{cycle-celeb}{b} shows the stochastic latent code of the network, where the input noise vector is held fixed across each column. We can see that the encoder network can learn stochastic image-to-image translation of female faces to male faces, and that the noise vector has captured the style of the male face images such as variations in eyeglasses, hair, mustache, or beard. \myfigg{cycle-celeb}{c} shows the deterministic reconstructions of the original image from the stochastic latent code. We can see that the deterministic decoder learns to invert the stochasticity of the encoder mapping, and generates almost perfect reconstructions of the input images. Note that if instead of the joint distribution matching cost between $(\mathbf{x},\hat{\mathbf{z}})$ and $(\hat{\mathbf{x}},\hat{\mathbf{z}})$, we use a marginal distribution matching cost between $\mathbf{x}$ and $\hat{\mathbf{x}}$ at the output, the decoder simply learns to generate a random female face, instead of reconstructing the original input female face.

\parhead{Related Works.} The problem of unsupervised multimodal image-to-image translation has been studied in several other works including Augmented-CycleGAN~\citep{aug-cyclegan}, MUNIT~\citep{munit} and DRIT~\citep{drit}. The Augmented-CycleGAN augments the CycleGAN's architecture with a style \emph{latent variable}; and MUNIT uses a domain-specific style latent variable, and a domain-invariant content code in its architecture. All these methods use a separate inference network to perform inference over the style latent variable. By augmenting the network architecture with the style code, the many-to-many mapping becomes a one-to-one mapping, and thus these methods can use different combinations of the CycleGAN's unimodal cycle-consistency and marginal distribution matching costs to learn stochastic mappings. The fundamental difference between CycleIAEs and these methods is that in the probabilistic model of CycleIAEs, the style \emph{noise vector} is not a latent variable, since we are marginalizing over the style noise vector to obtain the implicit decoder distribution. Therefore, we no longer need to perform inference for the style, and we only have a single inference network that performs inference over the content latent variable $\mathbf{z}$.

\section{Flipped Implicit Autoencoders}\label{sec:fiae}

\begin{figure}[t]
\begin{center}
\includegraphics[scale=0.5]{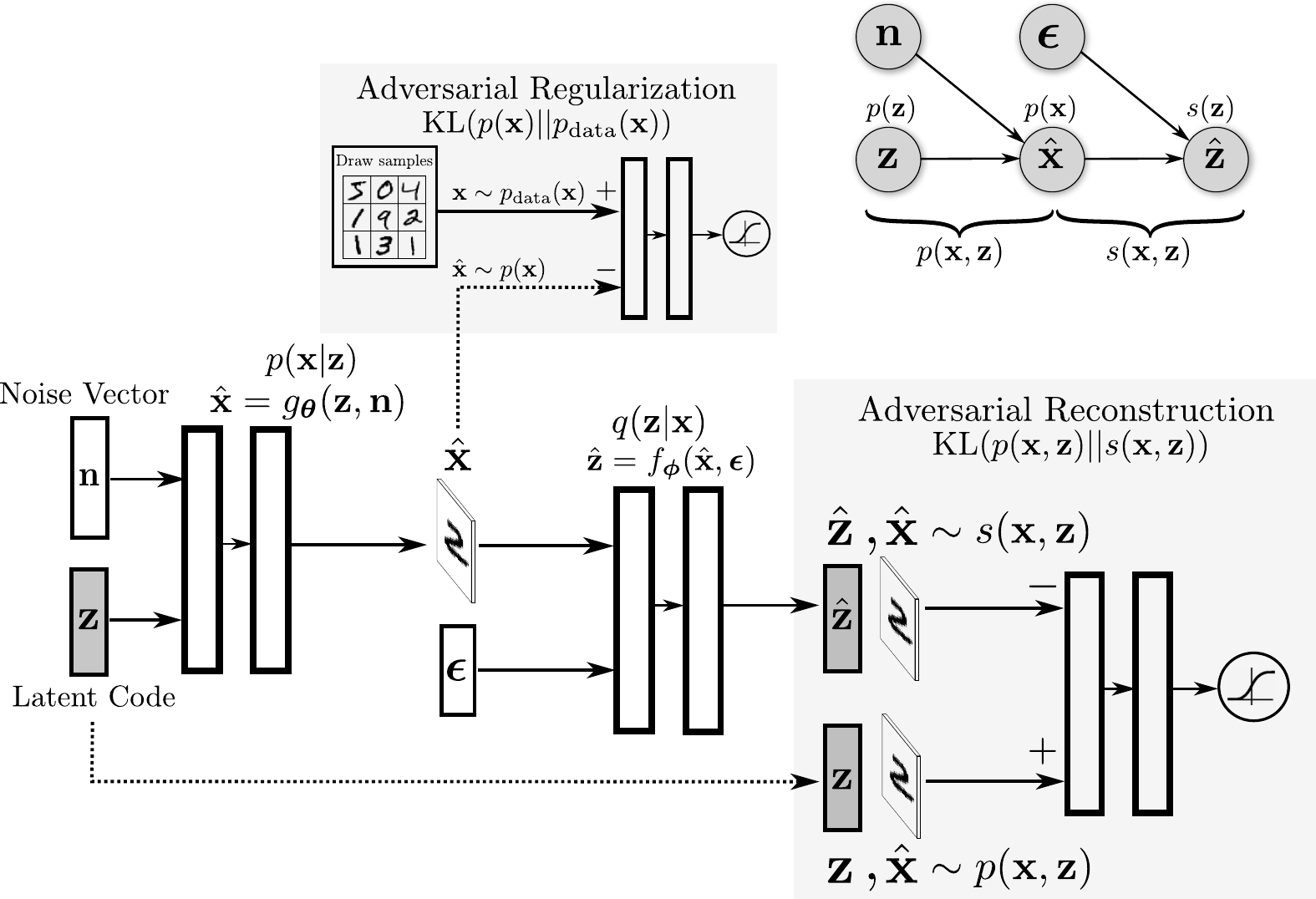}
\end{center}
\vspace{-.3cm}
\caption{\label{fig:fiae}Architecture and graphical model of flipped implicit autoencoders.}
\end{figure}

In this section, we describe a variant of IAEs called ``Flipped Implicit Autoencoder'' (FIAE), and show its applications in learning expressive variational inference networks for GANs.
Let $\mathbf{z}$ be the latent code that comes from the prior distribution $p(\mathbf{z})$. The encoder of the FIAE (\myfig{fiae}) parametrizes an implicit distribution that uses the noise vector $\mathbf{n}$ to define the conditional likelihood distribution $p(\mathbf{x}|\mathbf{z})$. The decoder of the FIAE parametrizes an implicit distribution that uses the noise vector $\bepsilon$ to define the variational posterior distribution $q(\mathbf{z}|\mathbf{x})$.
In addition to the distributions defined in \mysec{iae}, we also define the \emph{joint latent reconstruction distribution} $s(\mathbf{x}, \mathbf{z})$, and the \emph{aggregated latent reconstruction distribution} $s(\mathbf{z})$ as follows:
\begin{align}\quad s(\mathbf{x}, \mathbf{z}) = p(\mathbf{x})q(\mathbf{z}|\mathbf{x}) \qquad \hat{\mathbf{z}}\sim s(\mathbf{z}) = \int_\mathbf{x} s(\mathbf{x}, \mathbf{z}) d\mathbf{x}\end{align}
The objective of the standard variational inference is minimizing $\text{KL}( q(\mathbf{x}, \mathbf{z})\|p(\mathbf{x},\mathbf{z}))$, which is the variational upper-bound on $\text{KL}( p_\text{data}(\mathbf{x})\|p(\mathbf{x}))$. The objective of FIAEs is the reverse KL divergence $\text{KL}( p(\mathbf{x}, \mathbf{z})\|q(\mathbf{x},\mathbf{z}))$, which is the variational upper-bound on $\text{KL}( p(\mathbf{x})\|p_\text{data}(\mathbf{x}))$. The FIAE optimizes this variational bound by splitting it into a reconstruction term and a regularization term as follow:
\begin{align}
\text{KL}( p(\mathbf{x})\|p_\text{data}(\mathbf{x}))
&\leq
\underbrace{\text{KL}( p(\mathbf{x}, \mathbf{z})\|q(\mathbf{x},\mathbf{z}))}_\text{Variational Bound} \\
&=
\underbrace{\text{KL}( p(\mathbf{x})\|p_\text{data}(\mathbf{x}))}_\text{InfoGAN Regularization}
+ \underbrace{\mathbb{E}_{\mathbf{z} \sim p(\mathbf{z})}\Big[\mathbb{E}_{p(\mathbf{x}|\mathbf{z})} [-\log q(\mathbf{z}|\mathbf{x})]\Big]}_\text{InfoGAN Reconstruction} 
-\underbrace{\mathcal{H}(\mathbf{z}|\mathbf{x})}_\text{Cond. Entropy} \label{eq:infogan} \\
&=
\underbrace{\text{KL}( p(\mathbf{x})\|p_\text{data}(\mathbf{x}))}_\text{FIAE Regularization} 
+ \underbrace{\mathbb{E}_{\mathbf{x} \sim p(\mathbf{x})}\Big[\text{KL}( p(\mathbf{z}|\mathbf{x})\|q(\mathbf{z}|\mathbf{x}))\Big]}_\text{FIAE Reconstruction} \label{eq:fiae1} \\
&=
\underbrace{\text{KL}( p(\mathbf{x})\|p_\text{data}(\mathbf{x}))}_\text{FIAE Regularization} 
+ \underbrace{\text{KL}(p(\mathbf{x},\mathbf{z})\|s(\mathbf{x},\mathbf{z}))}_\text{FIAE Reconstruction} \label{eq:fiae2}
\end{align}
\noindent
where the conditional entropy $\mathcal{H}(\mathbf{z}|\mathbf{x})$ is defined under the joint model distribution $p(\mathbf{x}, \mathbf{z})$. Similar to IAEs, the FIAE has a regularization term and a reconstruction term (\myeq{fiae1} and \myeq{fiae2}). The regularization cost uses a GAN to train the encoder (conditional likelihood) such that the model distribution $p(\mathbf{x})$ matches the data distribution $p_\text{data}(\mathbf{x})$. The reconstruction cost uses a GAN to train both the encoder (conditional likelihood) and the decoder (variational posterior) such that the joint model distribution $p(\mathbf{x},\mathbf{z})$ matches the joint latent reconstruction distribution $s(\mathbf{x},\mathbf{z})$.

\parhead{Connections with ALI and BiGAN.} In ALI~\citep{ali} and BiGAN~\citep{bigan} models, the input to the recognition network is the samples of the real data $p_\text{data}(\mathbf{x})$; however, in FIAEs, the recognition network only gets to see the synthetic samples that come from the simulated data $p(\mathbf{x})$, while at the same time, the regularization cost ensures that the simulated data distribution is close the real data distribution. Training the recognition network on the simulated data in FIAEs is in spirit similar to the ``sleep'' phase of the wake-sleep algorithm~\citep{wake_sleep}, during which the recognition network is trained on the samples that the network ``dreams'' up. One of the flaws of training the recognition network on the simulated data is that early in the training, the simulated data do not look like the real data, and thus the recognition path learns to invert the generative path in part of the data space that is far from the real data distribution. As the result, the reconstruction GAN might not be able to keep up with the moving simulated data distribution and get stuck in a local optimum. However, in our experiments with FIAEs, we did not find this to be a major problem.

\begin{figure}[t]
\begin{minipage}{\linewidth}
    \centering
    \begin{minipage}{0.49\linewidth}
        \begin{figure}[H]
          \subfigure[True Posterior]{
          \includegraphics[scale=0.1]{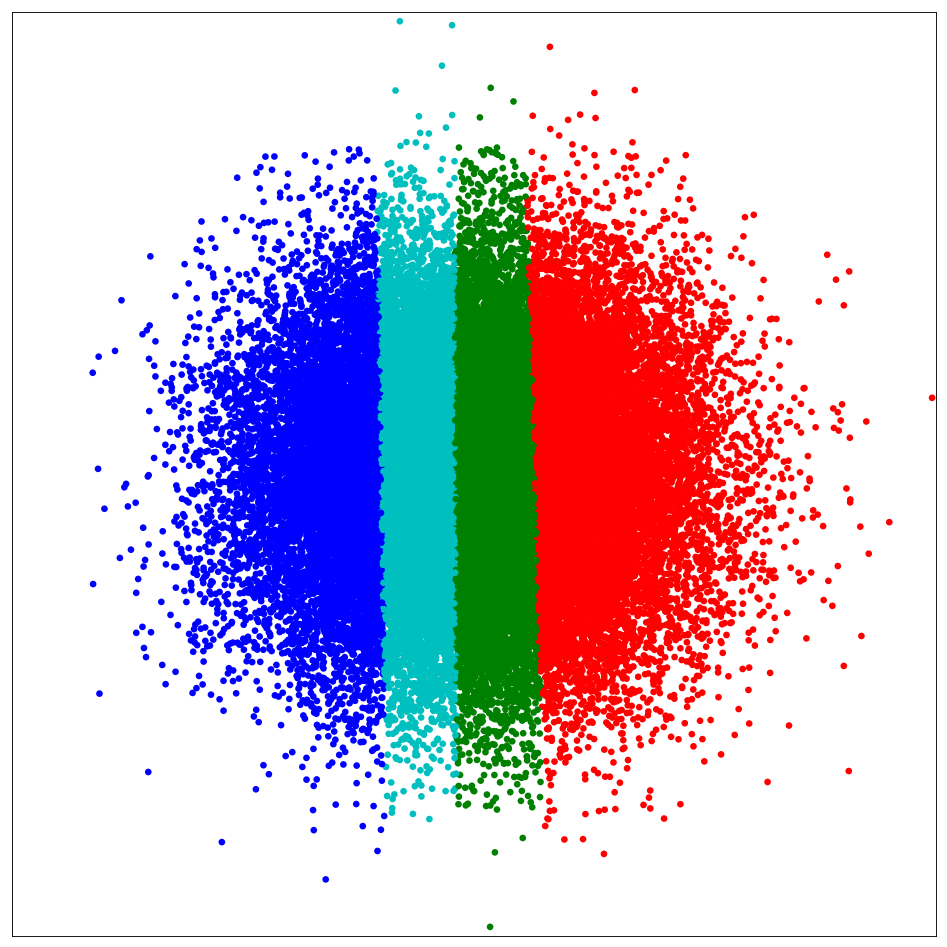}}
          \hspace{-.2cm}          
          \subfigure[Variational Posterior]{
          \includegraphics[scale=0.1]{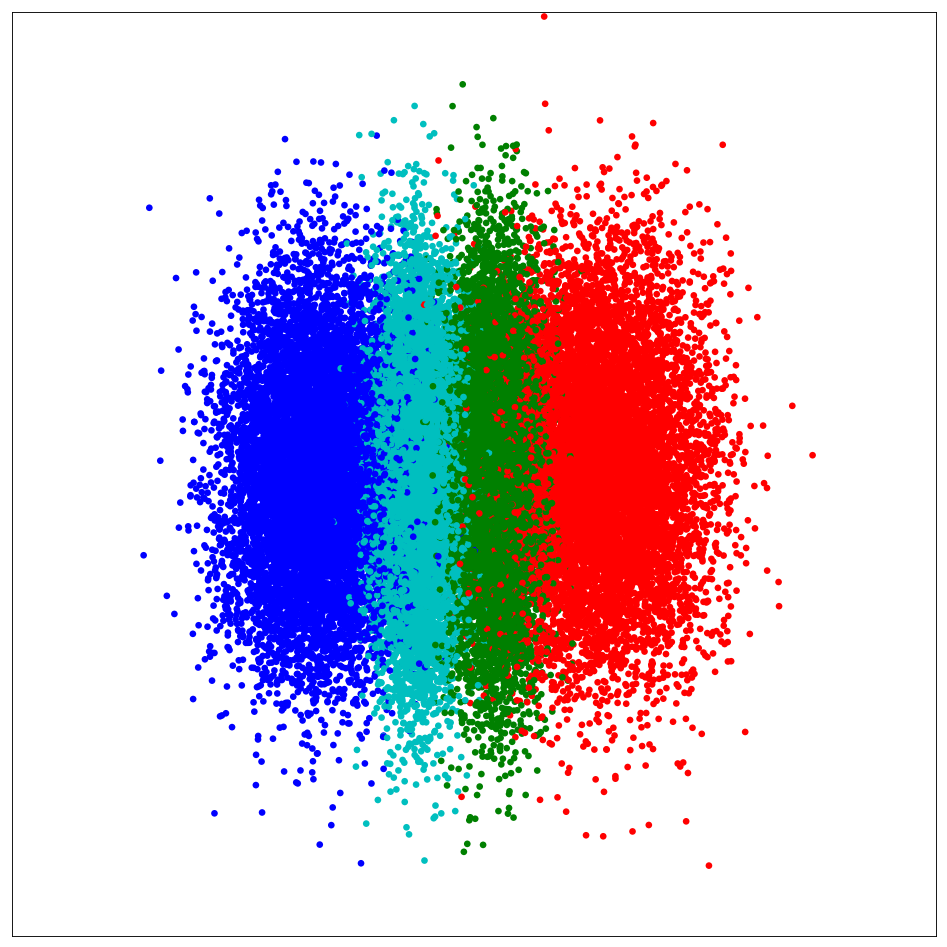}}
          \hspace{-.2cm}
          \vspace{-.3cm}
          \caption{\label{fig:toy_infogan}InfoGAN on a toy dataset. (a) True posterior. (b) Factorized Gaussian variational posterior.}
        \end{figure}
    \end{minipage}
    \hspace{.1cm}
    \begin{minipage}{0.49\linewidth}
        \begin{figure}[H]
          \subfigure[True Posterior]{
          \includegraphics[scale=0.1]{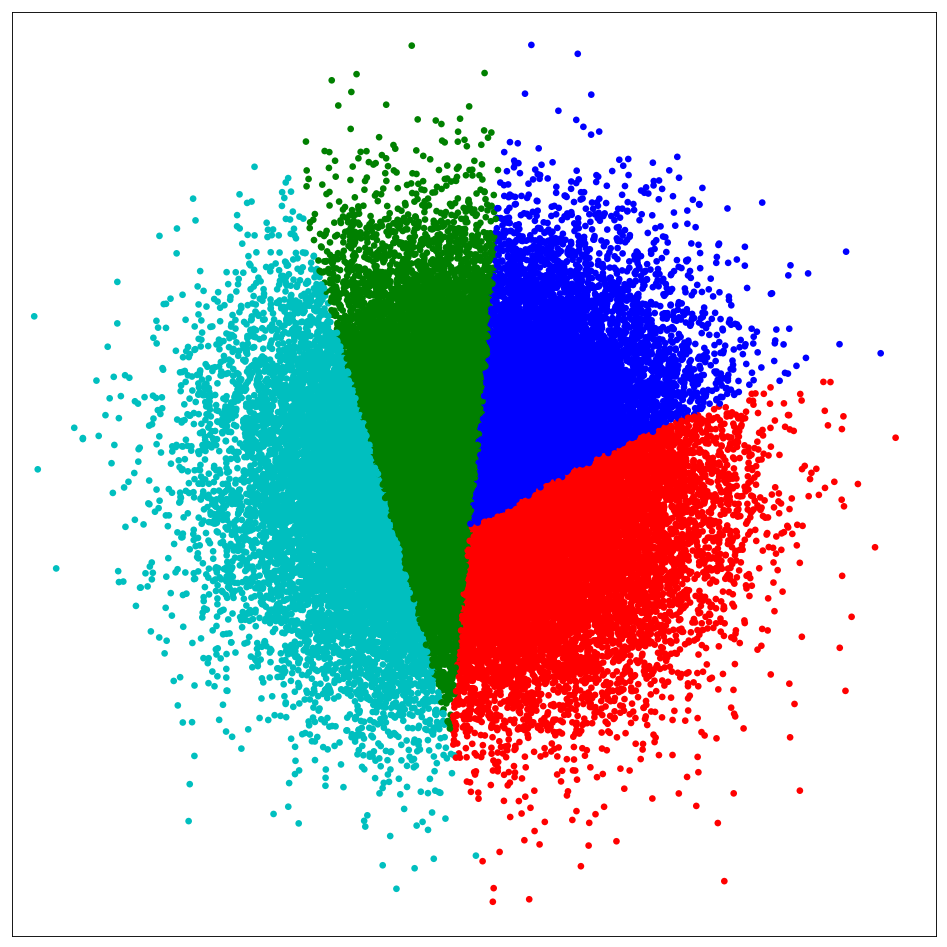}}
          \hspace{-.2cm}          
          \subfigure[Variational Posterior]{
          \includegraphics[scale=0.1]{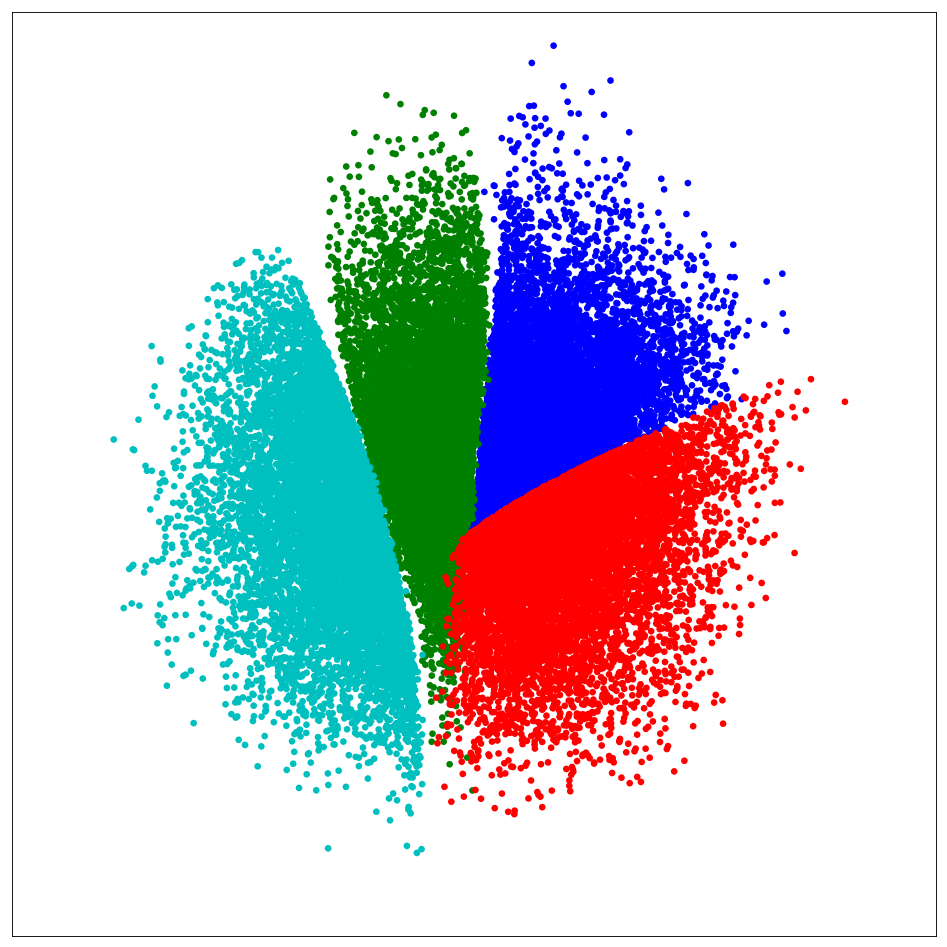}}
          \hspace{-.2cm}   
          \vspace{-.3cm}       
          \caption{\label{fig:toy_fiae}Flipped implicit autoencoder on a toy dataset. (a) True posterior. (b) Implicit variational posterior.}
          \end{figure}
    \end{minipage}
\end{minipage}
\end{figure}

\parhead{Connections with InfoGAN.} InfoGANs~\citep{infogan}, similar to FIAEs, train the variational posterior network on the simulated data; however, as shown in \myeq{infogan}, InfoGANs use an explicit reconstruction cost function (e.g., Euclidean cost) on the code space for learning the variational posterior. In order to compare FIAEs and InfoGANs, we train them on a toy dataset with four data-points and use a 2D Gaussian prior (\myfig{toy_infogan} and \myfig{toy_fiae}). Each colored cluster corresponds to the posterior distribution of one data-point. In InfoGANs, using the Euclidean cost to reconstruct the code is equivalent to learning a factorized Gaussian variational posterior distribution (\myfigg{toy_infogan}{b})\footnote{In \myfigg{toy_infogan}{b}, we have trained both the mean and the standard deviation of the Gaussian posteriors.}. This constraint on the variational posterior restricts the family of the conditional likelihoods that the model can learn by enforcing the generative path to learn a conditional likelihood whose true posterior could fit to the factorized Gaussian approximation of the posterior. For example, we can see in \myfigg{toy_infogan}{a} that the model has learned a conditional likelihood whose true posterior is axis-aligned, so that it could better match the factorized Gaussian variational posterior (\myfigg{toy_infogan}{b}). In contrast, the FIAE can learn an arbitrarily expressive variational posterior distribution (\myfigg{toy_fiae}{b}), which enables the generative path to learn a more expressive conditional likelihood and true posterior (\myfigg{toy_fiae}{a}).

One of the main flaws of optimizing the reverse KL divergence is that the variational posterior will have the mode-covering behavior rather than the mode-picking behavior.
For example, we can see from \myfigg{toy_infogan}{b} that the Gaussian posteriors of different data-points in InfoGAN have some overlap; but this is less of a problem in the FIAE (\myfigg{toy_fiae}{b}), as it can learn a more expressive $q(\mathbf{z}|\mathbf{x})$. This mode-averaging behavior of the posterior can be also observed in the wake-sleep algorithm, in which during the sleep phase, the recognition network is trained using the reverse KL divergence objective.

The FIAE objective is not only an upper-bound on $\text{KL}( p(\mathbf{x})\|p_\text{data}(\mathbf{x}))$, but is also an upper-bound on $\text{KL}( p(\mathbf{z})\|q(\mathbf{z}))$ and $\text{KL}( p(\mathbf{z}|\mathbf{x})\|q(\mathbf{z}|\mathbf{x}))$. As a result, the FIAE matches the variational posterior $q(\mathbf{z}|\mathbf{x})$ to the true posterior $p(\mathbf{z}|\mathbf{x})$, and also matches the aggregated posterior $q(\mathbf{z})$ to the prior $p(\mathbf{z})$. For example, we can see in \myfigg{toy_fiae}{b} that $q(\mathbf{z})$ is very close to the Gaussian prior. However, the InfoGAN objective is theoretically not an upper-bound on $\text{KL}( p(\mathbf{x})\|p_\text{data}(\mathbf{x}))$, $\text{KL}( p(\mathbf{z})\|q(\mathbf{z}))$ or $\text{KL}( p(\mathbf{z}|\mathbf{x})\|q(\mathbf{z}|\mathbf{x}))$. As a result, in InfoGANs, the variational posterior $q(\mathbf{z}|\mathbf{x})$ need not be close to the true posterior $p(\mathbf{z}|\mathbf{x})$, or the aggregated posterior $q(\mathbf{z})$ does not have to match the prior $p(\mathbf{z})$.

\begin{figure}[t]
\centering
\includegraphics[scale=.22]{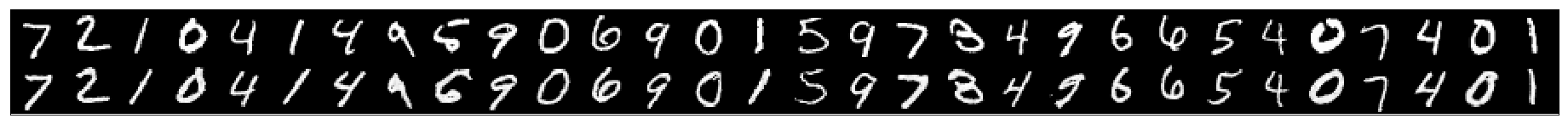}
\vspace{-.3cm}
\caption{\label{fig:mnist_fiae} Reconstructions of the flipped implicit autoencoder on the MNIST dataset. Top row shows the MNIST test images, and bottom row shows the deterministic reconstructions.}
\end{figure}

\subsection{Experiments of Flipped Implicit Autoencoders}
\parhead{Reconstruction.} In this section, we show that the variational posterior distribution of the FIAE can invert its conditional likelihood function by showing that the network can perform reconstructions of the images. We make both the conditional likelihood and the variational posterior deterministic by removing both noise vectors $\mathbf{n}$ and $\bepsilon$. \myfig{mnist_fiae} shows the performance of the FIAE with a code size of 15 on the test images of the MNIST dataset. The reconstructions are obtained by first passing the image through the recognition network to infer its latent code, and then using the inferred latent code at the input of the conditional likelihood to generate the reconstructed image.

\parhead{Clustering.}
Similar to IAEs, we can use FIAEs for clustering. We perform an experiment on the MNIST dataset by choosing a discrete categorical latent code $\mathbf{z}$ of size 10, which captures the digit identity; and a continuous Gaussian noise vector $\mathbf{n}$ of size 10, which captures the style of the digit. The variational posterior distribution $q(\mathbf{z}|\mathbf{x})$ is also parametrized by an implicit distribution with a Gaussian noise vector $\bepsilon$ of size 20, and performs inference only over the digit identity $\mathbf{z}$. Once the network is trained, we can use the variational posterior to cluster the test images of the MNIST dataset. This network achieves the error rate of about 2\% in classifying digits in an unsupervised fashion by matching each categorical code to a digit type. We observed that when there is uncertainty in the digit identity, different draws of the noise vector $\bepsilon$ results in different one-hot vectors at the output of the recognition network, showing that the implicit decoder can efficiently capture the uncertainty.

\section{Conclusion}
In this paper, we proposed the implicit autoencoder, which is a generative autoencoder that uses implicit distributions to learn expressive variational posterior and conditional likelihood distributions. We showed that in IAEs, the information of the data distribution is decomposed between the prior and the conditional likelihood. When using a low dimensional Gaussian distribution for the latent code, we showed that the IAE can disentangle high-level and abstract information from the low-level statistics. We also showed that by using a categorical latent code, we can learn discrete factors of variation and perform clustering and semi-supervised learning.
We further proposed cycle implicit autoencoders and showed that they can learn multimodal image-to-image mappings.
Finally, we proposed flipped implicit autoencoders and showed that they can learn expressive variational inference networks for GANs.

\bibliographystyle{unsrt}
{\small \bibliography{iae}}

\clearpage

\begin{appendices}

\section{Derivation of the ELBO of Implicit Autoencoders}\label{appendix:proof}

\begin{scriptsize}
\begin{align}
\mathbb{E}_{\mathbf{x} \sim p_d(\mathbf{x})}[\log p(\mathbf{x})] &\geq
\mathbb{E}_{\mathbf{x} \sim p_d(\mathbf{x})} \Big[\mathbb{E}_{q(\mathbf{z}|\mathbf{x})} \log {p(\mathbf{x},\mathbf{z})\over q(\mathbf{z}|\mathbf{x})}\Big]
\\
&=
\int q(\mathbf{x},\mathbf{z})\log {p(\mathbf{x},\mathbf{z})\over q(\mathbf{z}|\mathbf{x})} d\mathbf{x}d\mathbf{z} 
\\
&=
-\int q(\mathbf{x},\mathbf{z})\log {q(\mathbf{z}|\mathbf{x}) \over p(\mathbf{x}|\mathbf{z})} d\mathbf{x}d\mathbf{z}
+\int q(\mathbf{z})\log p(\mathbf{z}) d\mathbf{z}
\\
&=
-\int q(\mathbf{x},\mathbf{z}) \log {q(\mathbf{z}|\mathbf{x}) \over p(\mathbf{x}|\mathbf{z})} d\mathbf{x}d\mathbf{z}
-\int p_d(\mathbf{x}) \log p_d(\mathbf{x}) d\mathbf{x} 
+\int q(\mathbf{z}) \log p(\mathbf{z}) d\mathbf{z}
+\int p_d(\mathbf{x})\log p_d(\mathbf{x}) d\mathbf{x} 
\\
&=
-\int q(\mathbf{x},\mathbf{z}) \log {{p_d(\mathbf{x}) q(\mathbf{z}|\mathbf{x})} \over p(\mathbf{x}|\mathbf{z})} d\mathbf{x}d\mathbf{z}
+\int q(\mathbf{z}) \log p(\mathbf{z}) d\mathbf{z}
-\mathcal{H}_{\text{data}}{(\mathbf{x})}
\\
&=
-\int q(\mathbf{x},\mathbf{z}) \log {{q(\mathbf{x},\mathbf{z})} \over p(\mathbf{x}|\mathbf{z})} d\mathbf{x}d\mathbf{z}
+\int q(\mathbf{z}) \log q(\mathbf{z}) d\mathbf{z}
-\int q(\mathbf{z}) \log {q(\mathbf{z}) \over p(\mathbf{z})} d\mathbf{z}
-\mathcal{H}_{\text{data}}{(\mathbf{x})}
\\
&=
-\int q(\mathbf{x},\mathbf{z}) \log {{q(\mathbf{x},\mathbf{z})} \over {q(\mathbf{z}) p(\mathbf{x}|\mathbf{z})}} d\mathbf{x}d\mathbf{z}
-\text{KL}( q(\mathbf{z})\|p(\mathbf{z}))
-\mathcal{H}_{\text{data}}{(\mathbf{x})}
\\
&=
-\text{KL}(q(\mathbf{x},\mathbf{z})\|r(\mathbf{x},\mathbf{z}))
-\text{KL}( q(\mathbf{z})\|p(\mathbf{z}))
-\mathcal{H}_{\text{data}}{(\mathbf{x})}
\end{align}
\end{scriptsize}

\vspace{-.3cm}
\section{Implementation Details}\label{appendix:implementation}
\vspace{-.1cm}

\subsection{Implementation Details of IAEs}
\parhead{Network Architectures.}
The regularization discriminator in all the experiments is a two-layer neural network, where each layer has 2000 hidden units with the ReLU activation function. The architecture of the encoder, the decoder and the reconstruction discriminator for each dataset is described in \mytable{mnist-hyper}, \mytable{svhn-hyper}, and \mytable{celeb-hyper}.

\vspace{-.1cm}
\parhead{Optimizing Adversarial Reconstruction.} In order to minimize the adversarial reconstruction cost, we need to match $r(\mathbf{x},\mathbf{z})$ to the target distribution $q(\mathbf{x},\mathbf{z})$, which changes during the training. In order to do so, we could back-propagate through both the positive examples $(\mathbf{x},\hat{\mathbf{z}})$ and the negative examples $(\hat{\mathbf{x}},\hat{\mathbf{z}})$. However, we empirically observed that by only back-propagating through negative examples, we will provide a more stable target distribution $q(\mathbf{x},\mathbf{z})$ for $r(\mathbf{x},\mathbf{z})$ to aim for, which results in a more stable training dynamic and better empirical performance.

\parhead{Latent Code Conditioning.}
There are two methods to implement how the reconstruction GAN conditions on the latent code.

\parhead{1. Location-Dependent Conditioning.}
Suppose the size of the first convolutional layer of the discriminator is \texttt{(batch, width, height, channels)}. We use a one layer neural network with $1000$ ReLU hidden units to transform the latent code of size \texttt{(batch, latent\char`_code\char`_size)} to a spatial tensor of size \texttt{(batch, width, height, 1)}. We then broadcast this tensor across the channel dimension to get a tensor of size \texttt{(batch, width, height, channels)}, and then add it to the first layer of the discriminator as an adaptive bias. In this method, the latent code has spatial and location-dependent information within the feature map. This is the method that we used in deterministic and stochastic reconstruction experiments.

\parhead{2. Location-Invariant Conditioning.}
Suppose the size of the first convolutional layer of the discriminator is \texttt{(batch, width, height, channels)}. 
We use a linear mapping to transform the latent code of size \texttt{(batch, latent\char`_code\char`_size)} to a tensor of size \texttt{(batch, channels)}. We then broadcast this tensor across the width and height dimensions, and then add it to the first layer of the discriminator as an adaptive bias. In this method, the latent code is encouraged to learn the high-level information that is location-invariant such as the class label information. We used this method in all the clustering and semi-supervised learning experiments.

\vspace{-.2cm}
\subsection{Implementation Details of CycleIAEs}
\vspace{-.1cm}
In all the image-to-image translation experiments, we used the ResNet image generator used in~\citep{cyclegan} for both the encoder and the decoder of the autoencoder. In order to incorporate the noise vector in the ResNet architecture, we used conditional batch normalization~\citep{cin,adain} in all the layers. More specifically, for each layer, the noise vector is passed through a linear layer to predict the $\gamma$ and $\beta$ parameters of the batch-normalization for that layer.

\begin{table}[t]
\vspace{-10cm}
\centering
\resizebox{14cm}{!}{
\begin{tabular}{l|l|l}
  \toprule
Encoder & Decoder & Disc. Reconstruction GAN \\
\midrule
$\mathbf{x} \in \mathbb{R}^{28 \times 28}$  & $\hat{\mathbf{z}} \in \mathbb{R}^{20}$  and $\mathbf{n} \in \mathbb{R}^{100}$ & $(\mathbf{x},\hat{\mathbf{z}})$ or $(\hat{\mathbf{x}},\hat{\mathbf{z}})$	\\
FC. $2000$ ReLU.  		&  FC. $1024$ ReLU. BN 							       &	$4 \times 4$ Conv. $64$ ReLU. Stride $2$. BN		\\
FC. $2000$ ReLU.        &  FC. $128 \times 7 \times 7$ ReLU. BN 			   &	$4 \times 4$ Conv. $128$ ReLU. Stride $2$. BN		\\
FC. $20$ Linear. BN  	&  $4 \times 4$ UpConv. $64$ ReLU. Stride $2$. BN 	   &	FC. $1024$ ReLU. BN								    \\
						&  $4 \times 4$ UpConv. $1$ Sigmoid. Stride $2$. 	   &	FC. $1$ Linear									    \\
\bottomrule
\end{tabular}
}
\caption{\label{table:mnist-hyper}MNIST Hyper-Parameters.}
\vspace{1cm}
\resizebox{14cm}{!}{
\begin{tabular}{l|l|l}
  \toprule
Encoder & Decoder & Disc. Reconstruction GAN \\
\midrule
$\mathbf{x} \in \mathbb{R}^{32 \times 32 \times 3}$  & $\hat{\mathbf{z}} \in \mathbb{R}^{75}$  and $\mathbf{n} \in \mathbb{R}^{1000}$ & $(\mathbf{x},\hat{\mathbf{z}})$ or $(\hat{\mathbf{x}},\hat{\mathbf{z}})$	\\
$4 \times 4$ Conv. $64$ ReLU. Stride $2$. BN   &  FC. $256 \times 4 \times 4$ ReLU. BN             &	$4 \times 4$ Conv. $64$ ReLU. Stride $2$. BN		\\
$4 \times 4$ Conv. $128$ ReLU. Stride $2$. BN  &  $4 \times 4$ UpConv. $128$ ReLU. Stride $2$. BN  &	$4 \times 4$ Conv. $128$ ReLU. Stride $2$. BN		\\
$4 \times 4$ Conv. $256$ ReLU. Stride $2$. BN  &  $4 \times 4$ UpConv. $64$ ReLU. Stride $2$. BN   &	$4 \times 4$ Conv. $256$ ReLU. Stride $2$. BN		\\
FC. $75$ Linear. BN	    		     		   &  $4 \times 4$ UpConv. $3$ Tanh. Stride $2$.       &	FC. $1$ Linear									    \\
\bottomrule
\end{tabular}
}
\caption{\label{table:svhn-hyper}SVHN Hyper-Parameters.}
\vspace{1cm}
\resizebox{14cm}{!}{
\begin{tabular}{l|l|l}
  \toprule
Encoder & Decoder & Disc. Reconstruction GAN \\
\midrule
$\mathbf{x} \in \mathbb{R}^{48 \times 48 \times 3}$  & $\hat{\mathbf{z}} \in \mathbb{R}^{50}$  and $\mathbf{n} \in \mathbb{R}^{1000}$ & $(\mathbf{x},\hat{\mathbf{z}})$ or $(\hat{\mathbf{x}},\hat{\mathbf{z}})$	\\
$6 \times 6$ Conv. $128$ ReLU. Stride $2$. BN  &  FC. $512 \times 6 \times 6$ ReLU. BN             &	$6 \times 6$ Conv. $128$ ReLU. Stride $2$. BN		\\
$6 \times 6$ Conv. $256$ ReLU. Stride $2$. BN  &  $6 \times 6$ UpConv. $256$ ReLU. Stride $2$. BN  &	$6 \times 6$ Conv. $256$ ReLU. Stride $2$. BN		\\
$6 \times 6$ Conv. $512$ ReLU. Stride $2$. BN  &  $6 \times 6$ UpConv. $128$ ReLU. Stride $2$. BN  &	$6 \times 6$ Conv. $512$ ReLU. Stride $2$. BN		\\
FC. $50$ Linear. BN	                           &  $6 \times 6$ UpConv. $3$ Tanh. Stride $2$.       &	FC. $1$ Linear		                                \\
\bottomrule
\end{tabular}
}
\caption{\label{table:celeb-hyper}CelebA Hyper-Parameters.}
\end{table}

\end{appendices}

\end{document}